\documentclass[10pt,twocolumn,letterpaper]{article}
\makeatletter

\usepackage{iccv}              % To produce the CAMERA-READY version
\definecolor{iccvblue}{rgb}{0.21,0.49,0.74}
\usepackage[pagebackref,breaklinks,colorlinks,allcolors=iccvblue]{hyperref}

\usepackage{amsmath}
\usepackage{enumerate}
\usepackage{pifont}
\usepackage[dvipsnames]{xcolor}
\definecolor{mygreen}{rgb}{0, 0.59, 0}
\definecolor{myblue}{rgb}{0.2, 0.4, 1}
\usepackage{multirow}
\usepackage{amsmath,amssymb}   % 数学公式支持
\usepackage{algorithm}         % 浮动体环境：algorithm
\usepackage{algpseudocode}     % 伪代码命令：algorithmicx + algpseudocode
\usepackage{bibunits}
\defaultbibliographystyle{ieeenat_fullname}
\defaultbibliography{main}

\usepackage{etoolbox} % 放在导言区
\usepackage[accsupp]{axessibility}  % Improves PDF readability for those with disabilities.

\def\paperID{12101} % *** Enter the Paper ID here
\def\confName{ICCV}
\def\confYear{2025}

\title{TARS: Traffic-Aware Radar Scene Flow Estimation}

\author{%
  Jialong Wu\textsuperscript{1,3}\quad
  Marco Braun\textsuperscript{3}\quad
  Dominic Spata\textsuperscript{3}\quad
  Matthias Rottmann\textsuperscript{1,2}\\[1ex]
  \textsuperscript{1}University of Wuppertal \quad \textsuperscript{2}Osnabrück University \quad
  \textsuperscript{3}Aptiv Services Deutschland GmbH\\[1ex]
}

\begin{document}
\maketitle

\begin{bibunit}
    \begin{abstract}
Scene flow provides crucial motion information for autonomous driving. Recent LiDAR scene flow models utilize the rigid-motion assumption at the instance level, assuming objects are rigid bodies. However, these instance-level methods are not suitable for sparse radar point clouds. In this work, we present a novel Traffic‑Aware Radar Scene-Flow (TARS) estimation method, which utilizes motion rigidity at the traffic level. To address the challenges in radar scene flow, we perform object detection and scene flow jointly and boost the latter. We incorporate the feature map from the object detector, trained with detection losses, to make radar scene flow aware of the environment and road users. From this, we construct a Traffic Vector Field (TVF) in the feature space to achieve holistic traffic-level scene understanding in our scene flow branch. When estimating the scene flow, we consider both point-level motion cues from point neighbors and traffic-level consistency of rigid motion within the space. TARS outperforms the state of the art on a proprietary dataset and the View-of-Delft dataset, improving the benchmarks by 23\% and 15\%, respectively.
\end{abstract}    
    \section{Introduction} \label{sec:intro}

Scene flow estimates displacement vectors that describe point motion between two point cloud frames, which facilitates subsequent decision making in autonomous driving.

Early point cloud scene flow methods \cite{FlowNet3d, PointPWC} typically extract point-level or point-patch features and then estimate the flow by aggregating neighboring information from two frames. Considering only point-level information can result in points from the same object moving in different directions with varying magnitudes \cite{ICP-Flow}. Recent LiDAR scene flow methods focus on the rigid-motion assumption \cite{WsRSF}: objects move rigidly without deformation and a scene can be viewed as multiple rigidly moving objects and stationary parts. These LiDAR-based methods often begin with foreground segmentation, followed by clustering and matching to obtain instance pairs of the same object in two frames \cite{PCAccumulation}. Then they predict the rigid motion at the instance level for each object pair \cite{RigidGRU} or derive the instance-wise transformation using optimization-based methods \cite{RigidFlow, RCP}.

\begin{figure}[t] \centering
    \includegraphics[width=0.47\textwidth]{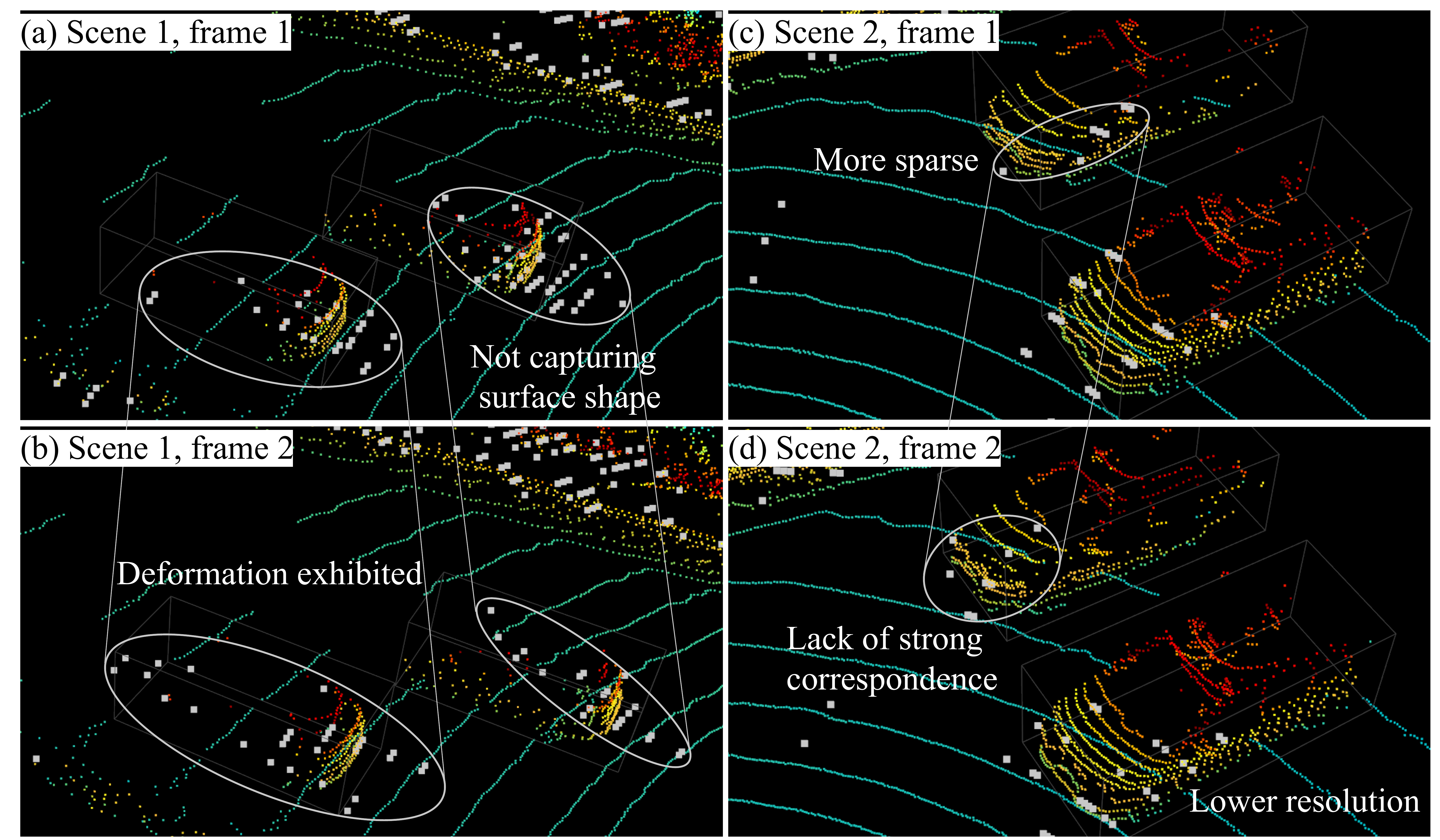}
    \caption{Challenges in radar scene flow. LiDAR points are shown in color, with corresponding radar points overlaid as larger gray points. (a-b) and (c-d) are two pairs of consecutive frames.} \label{fig:sec1_intro}
\end{figure}

However, these instance-level methods are not suitable for radar scene flow due to the inherent sparsity of radar data (cf.~\cref{fig:sec1_intro}). Radar is more robust under different weather conditions and is typically an order of magnitude less expensive than LiDAR \cite{review}. Nevertheless, radar point clouds are considerably sparser than LiDAR ones and fail to capture object shapes. These challenges lead to a lack of reliable correspondences for matching instance pairs. Objects in radar point clouds may even exhibit deformations between frames due to the sparsity. Inferring rigid motion at the instance level may misinterpret such abnormal deformations as motion. Moreover, even nearby objects may have only a few reflection points (Fig.\ \hyperref[fig:sec1_intro]{1d}), making optimization-based methods \cite{RigidFlow, ICP-Flow} ineffective.

In this work, we focus on reconciling the contradiction between the rigid-motion assumption and the sparsity of radar point clouds. Although rigid motion of objects under radar is difficult to capture at the instance level, we believe this motion rigidity still exists within the space occupied by objects. Therefore, we aim for a higher-level scene understanding, beyond the instance level, focusing on the traffic level to capture rigid motion hidden in the traffic context. When estimating the scene flow, we not only consider the motion cues propagated from neighboring points, but also the consistency of spatial context. This still adheres to the rigid-motion assumption. In fact, previous methods also assume that motion rigidity exists in Euclidean space when performing clustering and pairing of LiDAR points \cite{WsRSF, ICP-Flow}. However, to address the aforementioned challenges in radar scene flow, we do not further refine this assumption down to the instance level.

On the other hand, perceiving the environment and road users in the traffic is beneficial for motion prediction. Moreover, scene flow complements object detection with crucial motion information, enabling a more comprehensive perception. Therefore, we perform scene flow estimation and object detection jointly. We provide traffic cues to scene flow estimation through the object detector's feature map, which has been trained with detection losses and contains all relevant features about road users and the environment.

In our network, we achieve the traffic-level scene understanding by building a traffic vector field (TVF). We define the TVF as: a discrete grid map that incorporates traffic information about road users and the environment, with each cell containing a vector representing the motion. A conceptual diagram is shown in Fig.\ \hyperref[fig:point-GRU]{5b}. Note that we do not define an explicit 2D vector field; instead, we embed this concept within the feature space. The traffic information is extracted from the feature map of the object detection branch, and the motion information is passed through the hierarchical architecture of our scene flow branch. In each level of the architecture, we extract point-level motion cues from point neighbors while also paying attention to traffic-level motion consistency in Euclidean space. We use a coarse-grid TVF to achieve a high-level scene understanding, rather than falling into point-level details.

We evaluate our model TARS on a proprietary dataset and the View-of-Delft (VOD) dataset \cite{VOD}. Quantitative results demonstrate that TARS exceeds the state-of-the-art (SOTA) scene-flow accuracy on these datasets by 23\% and 15\%, respectively. Radar scene flow heavily relies on radar sensors’ ability to measure object velocities. However, radar can only measure radial velocity, leading to significant underestimation of tangential motion. Qualitative results show that TARS effectively captures objects’ rigid motion while mitigating the tangential motion challenge.

Our main contributions are summarized as follows:
\begin{itemize}
    \item We present TARS, the traffic-aware radar scene flow model that addresses the challenges of radar by leveraging the object detection feature map to obtain traffic context.
    \item We design the traffic vector field encoder and decoder modules to encode traffic-level motion understanding into the TVF and capture rigid motion in Euclidean space.
    \item TARS achieves SOTA accuracy by a large margin on both the proprietary dataset and VOD dataset.
\end{itemize}
    \section{Related Work}
\begin{figure*}[ht] \centering

    \includegraphics[width=\textwidth]{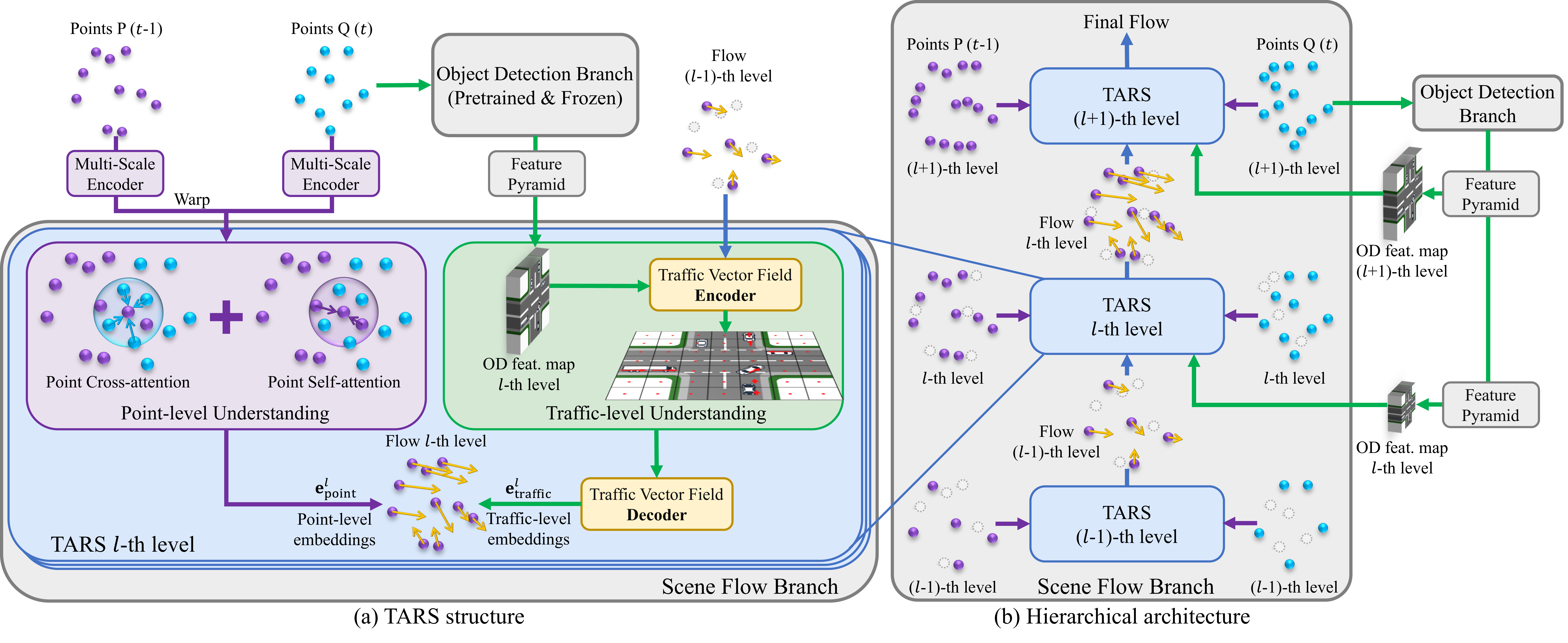}
    \caption{Overview of TARS. TARS employs a hierarchical architecture. At each level, it infers point-level motion cues using a double attention mechanism, while the TVF encoder leverages the OD feature map to build a traffic-level motion representation. The TVF decoder then extracts rigid motion cues in Euclidean space. Finally, dual-level flow embeddings are combined to estimate the scene flow.} \label{fig:sec3_archi}
    \vspace{-0.3cm}
\end{figure*}

\textbf{Point cloud scene flow.} Scene flow for LiDAR point clouds has been widely studied over the past few years. Early methods \cite{PointPWC, FlowNet3d, FlowStep3D} use PointNet \cite{PointNet++} as the point or patch feature extractor and calculate flow embeddings based on neighborhood information. Bi-PointFlowNet \cite{Bi-PointFlowNet} uses the bidirectional flow embeddings to capture the motion context between frames. PV-RAFT \cite{PV-RAFT} extracts point and voxel features to capture local and long-range correspondences. HALFlow \cite{HALFlow} applies a double attention mechanism to aggregate information from neighbors. DeFlow \cite{DeFlow} employs a gated recurrent unit (GRU) to transfer voxel features to points and improves efficiency on large-scale point clouds. Flow4D \cite{Flow4D} fuses multiple frames into 4D voxels and extracts spatio-temporal features. These methods infer scene flow only at the point-patch level, without considering higher-level motion consistency.

WsRSF \cite{WsRSF} and PCAccumulation \cite{PCAccumulation} utilize the aforementioned rigid-motion assumption at the instance level. They segment instance pairs and then regress the motion between each pair. Based on this approach, \citet{RigidGRU} introduce the nearest neighbor error minimization into a GRU to iteratively update the scene flow. Meanwhile, RigidFlow \cite{RigidFlow} uses the network output as the initialization for ICP (Iterative Closest Point) and refines the results. SCOOP \cite{SCOOP} trains a pure correspondence model and computes scene flow via optimization with smoothness prior. Let-It-Flow \cite{Let-It-Flow} enforces rigidity in object clustering. MBNSF \cite{MBNSF} encourages multi-body rigidity by adding a regularization term to the Chamfer loss. ICP-Flow \cite{ICP-Flow} is a non-learning method using clustering to obtain instance pairs and a histogram-based approach to initialize ICP. Although effective for LiDAR, these instance-level rigidity methods cannot handle the challenges in radar scene flow (\cref{fig:sec1_intro}).

\noindent \textbf{Radar scene flow.} Radar point cloud scene flow has only gained attention in recent years. MilliFlow \cite{milliFlow} estimates human motion. RaFlow \cite{RaFlow} employs motion segmentation and applies the Kabsch algorithm \cite{Kabsh} to estimate the ego-motion transformation as the static flow. It treats static points collectively, but lacks a higher-level understanding of dynamic points. CMFlow \cite{CMFlow} leverages additional cross-modal information as supervision signals to enhance the performance of radar scene flow. In this work, we provide additional traffic information for radar scene flow and capture the rigid motion of objects at the traffic level.

\noindent \textbf{Joint scene flow estimation \& object detection.} Combining these two tasks enables comprehensive perception in autonomous driving. \citet{SSLDet} train a shared backbone with alternating task-specific heads, while PillarFlowNet \cite{PillarFlowNet} uses a voxel representation and a multi-task head to jointly estimate scene flow and detect objects. PointFlowNet \cite{PointFlowNet} further infers point-wise motion from voxels. These fine-grained voxel-based methods are limited to point-patch-level motion understanding, with task interaction primarily via sharing backbone features. In contrast, our work provides a holistic traffic-level understanding for scene flow through synergistic interactions in the feature space. TrackFlow \cite{TrackFlow} directly derives LiDAR scene flow from object detection and tracking results. However, radar object tracking is significantly less accurate. Our approach perceives traffic context by leveraging feature maps, which reduces the reliance on object detection accuracy.

    \section{TARS Architecture} \label{sec:method}
We introduce TARS in a top-down approach. First, we present the hierarchical architecture that progressively refines scene flow (\cref{sec:3.1}). Next, we use the $l$-th level to explain the structure of TARS and its dual-level motion understanding (\cref{sec:3.2}). Thereafter, we dive into details of our TVF encoder, which encodes traffic and motion context into the TVF, achieving a traffic-level motion understanding (\cref{sec:3.3}). The TVF decoder captures rigid motion hidden in the surrounding context, and the scene flow head combines point and traffic-level flow embeddings to predict the scene flow (\cref{sec:3.4}). Finally, we briefly describe our recurrent module that leverages temporal cues (\cref{sec:3.5}).

\subsection{Hierarchical Architecture} \label{sec:3.1}

Following the hierarchical architecture in prior works \cite{PointPWC, HALFlow}, TARS has $L$ levels and predicts scene flow in a coarse-to-fine fashion, progressively refining the prediction. Perceiving the environment and road users in the traffic is beneficial for motion prediction, as they provide a traffic-level motion prior. The object detection (OD) feature map, trained with detection losses, contains all relevant features about road users and the environment. Therefore, we jointly perform scene flow estimation and object detection (see Fig.\ \hyperref[fig:sec3_archi]{2b}). We focus on enhancing the performance of radar scene flow, while the OD branch can be any detector, as long as it can generate bird's-eye view feature maps to provide traffic information for the scene flow branch.

The input to our scene flow branch consists of two point clouds $P \in \mathbb{R}^{N \times (3+2)}$ and $Q \in \mathbb{R}^{M \times (3+2)}$, with 5D initial features: $x,y,z$ coordinates plus RRV (relative radial velocity) and RCS (radar cross-section) \cite{review2}. The multi-scale point encoder \cite{PointNet++} is first applied to both point clouds for point feature extraction. Farthest point sampling is also performed to downsample the point clouds, yielding input point set pairs for each hierarchical level $\{ P^l \in \mathbb{R}^{N_l \times (3+C)}, Q^l  \in \mathbb{R}^{M_l \times (3+C)} \}^{L}_{l=1}$, where $C$ is the dimension of extracted point feature, $N_l$ and $M_l$ are downsampled by a factor $\gamma$. Then, starting from the smallest point set, the ($l$-1)-th level of TARS computes flow embeddings $\textbf{e}^{l-1} \in \mathbb{R}^{N_{l-1} \times D}$ and generates a coarse scene flow $F^{l-1} \in \mathbb{R}^{N_{l-1} \times 3}$, which are then used as input for the next level. After refining the flow using multiple TARS levels, we obtain the final scene flow for the full point set. 

From the second-lowest level, we enhance scene flow estimation by incorporating traffic information from the OD branch (green arrows in Fig.\ \hyperref[fig:sec3_archi]{2b}). Here, we feed the point cloud $Q$ to the OD branch. The reason for using the feature map from $Q$ rather than $P$ is that: in our hierarchical architecture, the point cloud $P$ is gradually warped toward the corresponding positions in $Q$ by each level's flow prediction. Therefore, using $Q$’s feature map allows for more accurate alignment between points and object features.

\subsection{TARS Structure} \label{sec:3.2}

In radar scene flow, capturing motion rigidity at the instance level is challenging (cf.~\cref{fig:sec1_intro}). However, we believe that rigid motion still exists regions occupied by objects. Therefore, with the help of the OD branch, we aim to achieve a traffic-level scene understanding to reveal the rigid motion hidden in the traffic context. Meanwhile, point-level matching information remains crucial for motion estimation. 
By integrating both point-level and traffic-level insights, TARS achieves a comprehensive motion understanding and enhances radar scene flow estimation.

Taking the $l$-th level as an example, Figure\ \hyperref[fig:sec3_archi]{2a} shows the structure of TARS. The point-level understanding extracts motion cues from neighboring points, while the traffic-level understanding is achieved by building a TVF (defined in \cref{sec:intro}) to capture the motion consistency. 

\subsubsection{Point‑Level Motion Understanding} \label{sec:3.1.1}

Point motion can be inferred from the matching information between neighboring points across consecutive point cloud frames \cite{PointPWC}. Previous studies \cite{FlowNet3d, PointPWC} use a multi-layer perceptron (MLP) to encode this point-level matching information, known as the cost volume, into the flow embeddings. However, we observed that these MLP-based methods are unstable in sparse radar point clouds due to larger point spacing and fewer reflections per object. Therefore, we use a double attention mechanism \cite{HALFlow} to adaptively extract the matching information from radar point clouds.

For a point $p_i^{l} \in \mathbb{R}^3$ in $P^l$, we compute cross-attention between $p_i^{l}$ and its $K$ nearest neighbors in $Q^l$ (blue circle in Fig.\ \hyperref[fig:sec3_archi]{2a}); then we apply self-attention between $p_i^{l}$ and its neighboring points in $P^l$ (purple circle in Fig.\ \hyperref[fig:sec3_archi]{2a}). Unlike HALFlow \cite{HALFlow}, we remove the direction vector to mitigate the point spacing issue, and we employ heterogeneous keys and values to obtain fully attentive flow embeddings.

Specifically, we first warp the point cloud $P^l$ closer to its neighbors in $Q^l$ using the upsampled coarse flow from the previous level: $P_{\textup{warp}}^l = P^l + \textup{Interp}(F^{l-1})$. For simplicity, we omit this in the equations below. Let $\textbf{p}_i^l, \textbf{q}_j^l \in \mathbb{R}^C$ represent the point features of  $p_i^l, q_j^l$. We compute the cross-attentive matching embeddings $\textbf{e}_{\textup{cross}}(p_i^l)$ for each point $p_i^l$:
\begin{flalign}
     & \textbf{e}_\textup{cross}(p_i^l) = \textup{Attention}(\textbf{p}_i^l, \mathcal{N}_Q(p_i^{l}), \mathcal{N}_Q(p_i^{l})), \\
     & \textup{Attention}(\cdot, \diamond, \star) = \textup{softmax}(\frac{\textup{Q}(\cdot) \textup{K}(\diamond)^T}{\sqrt{d_k}})\textup{V}(\star),
\end{flalign}
where $\mathcal{N}_Q(p_i^{l})=\textup{KNN}(p_i^{l}, Q^l)$ denotes the $K$ nearest neighbors of $p_i^{l}$ in $Q^l$. $\textup{Q}(\cdot), \textup{K}(\cdot), \textup{V}(\cdot)$ are linear layers.

Then, the point-level flow embeddings $\textbf{e}_{\textup{point}}(p_i^l)$ for a point $p_i^l$ is computed via self-attention as:
\begin{equation} \label{eq:e_point}
     \textbf{e}_\textup{point}(p_i^l) = \textup{Attention}(\textbf{e}_\textup{cross}(p_i^l), \mathcal{N}_{\textbf{e}}(p_i^{l}), \mathcal{N}_{\textbf{e}}(p_i^{l})),
\end{equation}
where $\mathcal{N}_{\textbf{e}}(p_i^{l})=\textup{KNN}(p_i^{l}, \textbf{e}_\textup{cross})$ fetches the matching embeddings for the neighbors of $p_i^{l}$ in $P^l$.

\subsubsection{Traffic‑Level Scene Understanding}

Our goal is to reconcile the rigid-motion assumption with the sparsity of radar point clouds. Instead of applying this assumption at the instance level \cite{WsRSF, ICP-Flow}, we capture the rigid motion at the traffic level to address the challenges in radar. To achieve this, we design a TVF encoder that builds a traffic-level scene understanding. Then we employ a TVF decoder to capture rigid motion hidden in the traffic context.

Specifically, the feature map $\chi_{\textup{od}}^l$ from the OD branch contains traffic information, and the flow embeddings $\textbf{e}^{l-1}$ passed from the previous level carry motion information. The TVF encoder combines and encodes them into a coarse-grid TVF in the feature space, enabling traffic-level scene understanding (green \& blue arrows in Fig.\ \hyperref[fig:sec3_archi]{2a}). The TVF decoder employs cross-attention between points and TVF grids to perceive rigid motion and generate the traffic-level flow embeddings. In the TVF encoder, we apply global attention to build a holistic traffic-level understanding within the TVF. In contrast, the TVF decoder restricts the cross-attention to a local area, which helps to capture the rigid motion in spatial context. Finally, we combine point-level $\textbf{e}_\textup{point}$ and traffic-level flow embeddings $\textbf{e}_\textup{traffic}$ to enhance scene flow estimation (purple \& green arrows in Fig.\ \hyperref[fig:sec3_archi]{2a}).

\subsection{Traffic Vector Field Encoder} \label{sec:3.3}

Our TVF encoder integrates traffic and motion information into the TVF, and it progressively updates this traffic-level motion representation across TARS's hierarchical levels.

\begin{figure}[tbp] \centering
    \includegraphics[width=0.48\textwidth]{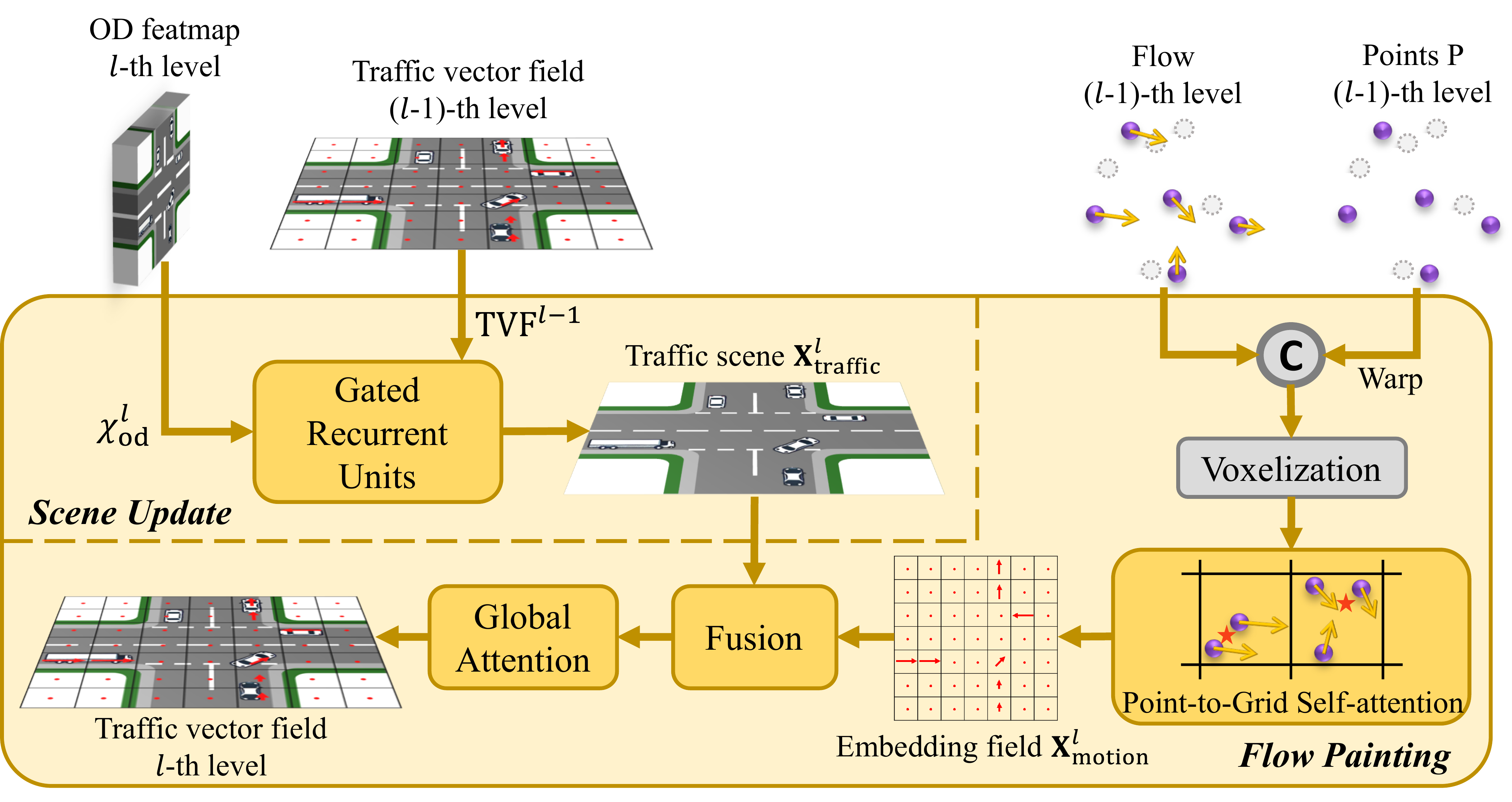}
    \caption{TVF encoder. Scene update: updates traffic information on the TVF using GRU; Flow painting: incorporates motion information into the TVF and build a holistic traffic representation.} \label{fig:encode}
    \vspace{-0.3cm}
\end{figure}

To this end, the TVF encoder employs two stages: \begin{enumerate*}[label=(\roman*)]
\item \textbf{Scene update}: A GRU \cite{ConvGRU} leverages the OD feature map to update the TVF from the previous level;
\item \textbf{Flow painting}: Point-to-grid self-attention adaptively paints the coarse flow from the previous level onto the scene representation.
\end{enumerate*}
Finally, we generate the new $\textup{TVF}^l \in \mathbb{R}^{ H \times W \times D_{\textup{TVF}}}$ using global attention.
Figure \ref{fig:encode} illustrates these two stages. The TVF maintains the same shape across each level and is configured to a coarse grid, \eg, $2\mathrm{m} \times 2\mathrm{m}$, enabling a high-level scene understanding without falling into point-level details.

\noindent \textbf{Scene update.} In this stage, we update the scene representation by leveraging traffic features from the $l$-th level of the OD feature pyramid to refine the previous level’s $\textup{TVF}^{l-1}$.

First, we apply CNN and pooling layers to the feature map $\chi_{\textup{od}}^l$ to adapt the object features and match the pre-defined shape of the coarse TVF. Then we use $\chi_{\textup{od}}^l$ as the input to a GRU, with $\textup{TVF}^{l-1}$ as the hidden state. The GRU is applied in an inter-level manner for updating the scene representation $\textbf{X}_\textup{traffic}^l$ across hierarchical levels:
\begin{flalign} \label{eq:GRU}
     &  \Tilde{\textbf{X}}_\textup{traffic}^l = \textup{tanh}(\textbf{W}_{\textup{G}} \ast \chi_{\textup{od}}^l + \textbf{U}_{\textup{G}} \ast (\textbf{r}^l \odot \textup{TVF}^{l-1})),  \\
     & \textbf{X}_\textup{traffic}^l = \textbf{z}^l \odot   \textup{TVF}^{l-1} +  (1- \textbf{z}^l) \odot   \Tilde{\textbf{X}}_\textup{traffic}^l,
\end{flalign}
where $\ast$ is the convolution operation, $\odot$ is the element-wise multiplication, $\textbf{W}_{\textup{G}}, \textbf{U}_{\textup{G}}$ are 2D convolution kernels,  $\textbf{r}^l, \textbf{z}^l$ are the reset and update gates, see \cite{ConvGRU} for details.

\noindent \textbf{Flow painting.} In this stage, we project the coarse flow from the previous level onto the grid, using point-to-grid self-attention. Next, we fuse the motion representation with the scene representation and apply global attention to build a high-level scene understanding.

Specifically, we concatenate the ($l$-1)-th level's flow embeddings $\textbf{e}^{l-1}$ with the point features $\textbf{p}^{l-1}$ extracted by the multi-scale encoder, and then project them onto the predefined 2D grid. Because our TVF grid is coarse, each cell may contain multiple points with varying motion patterns (see \cref{fig:encode}). Therefore, we apply point-to-grid self-attention to adaptively extract motion features. We perform both channel-wise and point-wise self-attention within each grid cell and obtain the motion embedding field $\textbf{X}_\textup{motion}^l$.

Next, we fuse the traffic feature $\textbf{X}_\textup{traffic}^l$ and motion feature $\textbf{X}_\textup{motion}^l$ using spatial attention \cite{SpatialAtt}. We concatenate the two feature maps and process them through CNN layers followed by a pixel-wise softmax to generate spatial attention weights. The fused traffic-level feature $\textbf{X}_\textup{fusion}^l$ is obtained as a weighted sum of $\textbf{X}_\textup{traffic}^l$ and $\textbf{X}_\textup{motion}^l$.

High-level scene understanding should not be limited to local areas. A global receptive field is crucial for modeling dependencies of rigid-body motions in traffic, \eg, the motion patterns of vehicles in the same lane. We use axial attention \cite{AxAtt} to provide the global vision and build the traffic-level motion understanding. It splits a standard attention block into separate row-wise and column-wise components, reducing complexity yet preserving global context. By stacking $\omega$ axial attention blocks, we enhance the fused features $\textbf{X}_\textup{fusion}^l$ and obtain the final $\textup{TVF}^l$.

\begin{figure}[tbp] \centering
    \includegraphics[width=0.38\textwidth]{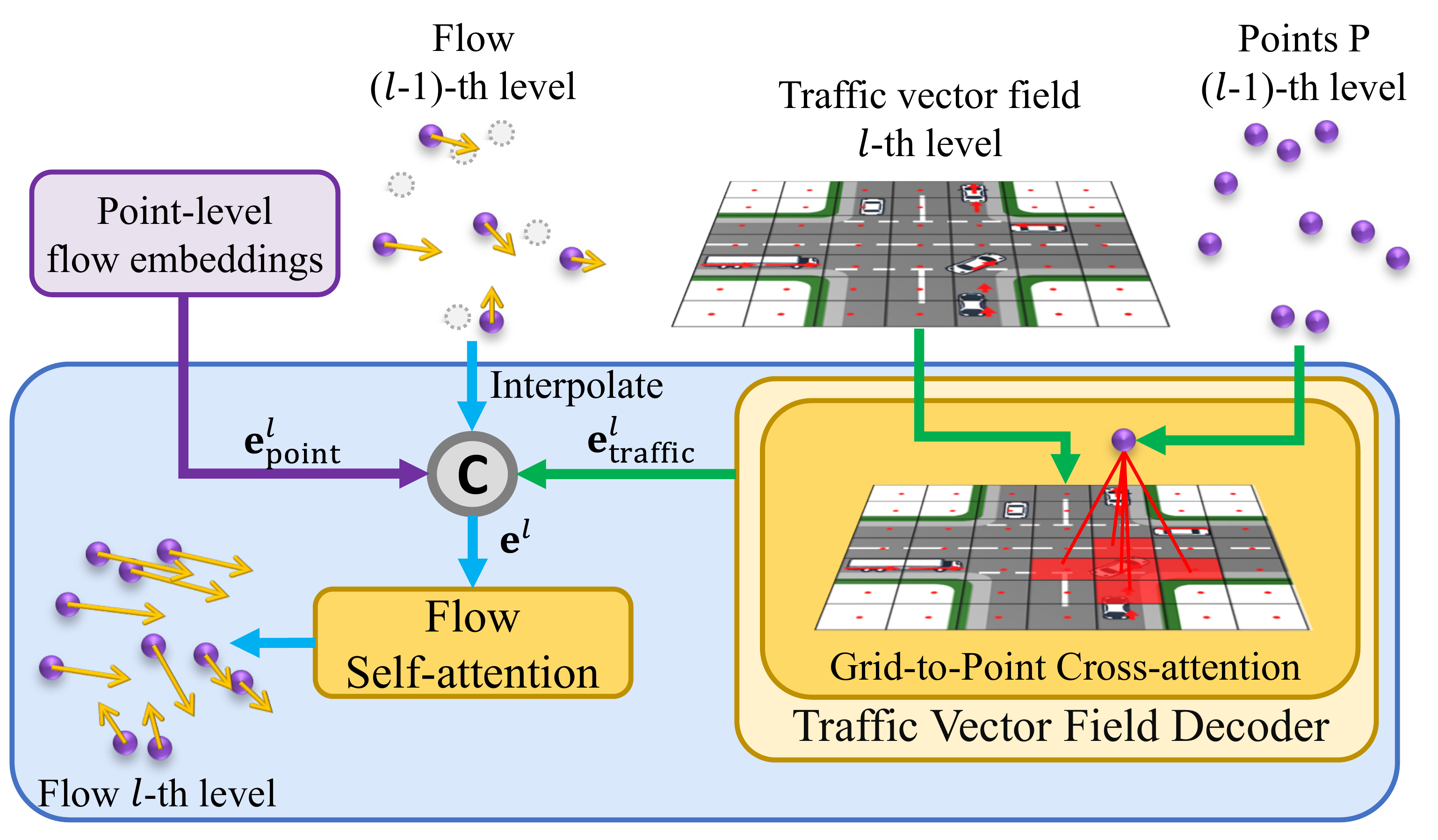}
    \caption{TVF decoder \& scene flow head: capture motion rigidity in spatial context; combine dual-level embeddings for prediction.} \label{fig:decode}
    \vspace{-0.3cm}
\end{figure}

\subsection{TVF Decoder \& Scene Flow Head} \label{sec:3.4}

Although radar point clouds are sparse, motion rigidity persists in Euclidean space. When building the traffic-level motion understanding, we encoded the rigid motion cues into the coarse-grid TVF. We then use the TVF decoder to perceive rigid motion hidden in the traffic context.

Specifically, we apply grid-to-point cross-attention between each point $p_i^{l}$ and its surrounding traffic context in the TVF, thereby integrating rigid motion cues into the flow embedding. To focus on relevant local rigid motion, the attentive receptive field is restricted to the nearby region around each point (see \cref{fig:decode}). We interpolate the coarse flow embeddings $\textbf{e}^{l-1}$ and concatenate with the point feature $\textbf{p}^l$ as the query, and the TVF grids as keys and values. This enables the resulting traffic-level flow embeddings $\textbf{e}_{\textup{traffic}}(p_i^l)$ to be aware of motion consistency in the traffic context:
\begin{flalign} \label{eq:decoder}
    & \hat{\textbf{p}}^l_i = \textup{Concat}(\textup{Interp}(\textbf{e}^{l-1})_i, \textbf{p}^l_i), \\ 
    & \textbf{e}_\textup{traffic}(p_i^l) = \textup{Attention}(\hat{\textbf{p}}^l_i, \mathcal{N}_{\textup{TVF}}(p_i^{l}), \mathcal{N}_{\textup{TVF}}(p_i^{l})),
\end{flalign}
where $\mathcal{N}_{\textup{TVF}}(p_i^{l})=\textup{KNN}(p_i^{l}, \textup{TVF}^l)$ fetches the surrounding $K$ cells of $p_i^{l}$ from $\textup{TVF}^l$.

Combining point-level and traffic-level motion understanding, we obtain the final flow embeddings $\textbf{e}^l = \textup{Concat}(\textbf{e}_\textup{point}, \textbf{e}_\textup{traffic}, \textup{Interp}(\textbf{e}^{l-1}))$. Finally, we apply another self-attention as in \cref{eq:e_point} on $\textbf{e}^l$, reduce it back to $C$ channels, and predict the final scene flow $F^l$.

\subsection{Temporal Update Module} \label{sec:3.5}
\begin{figure}[bp] \centering
    \includegraphics[width=0.45\textwidth]{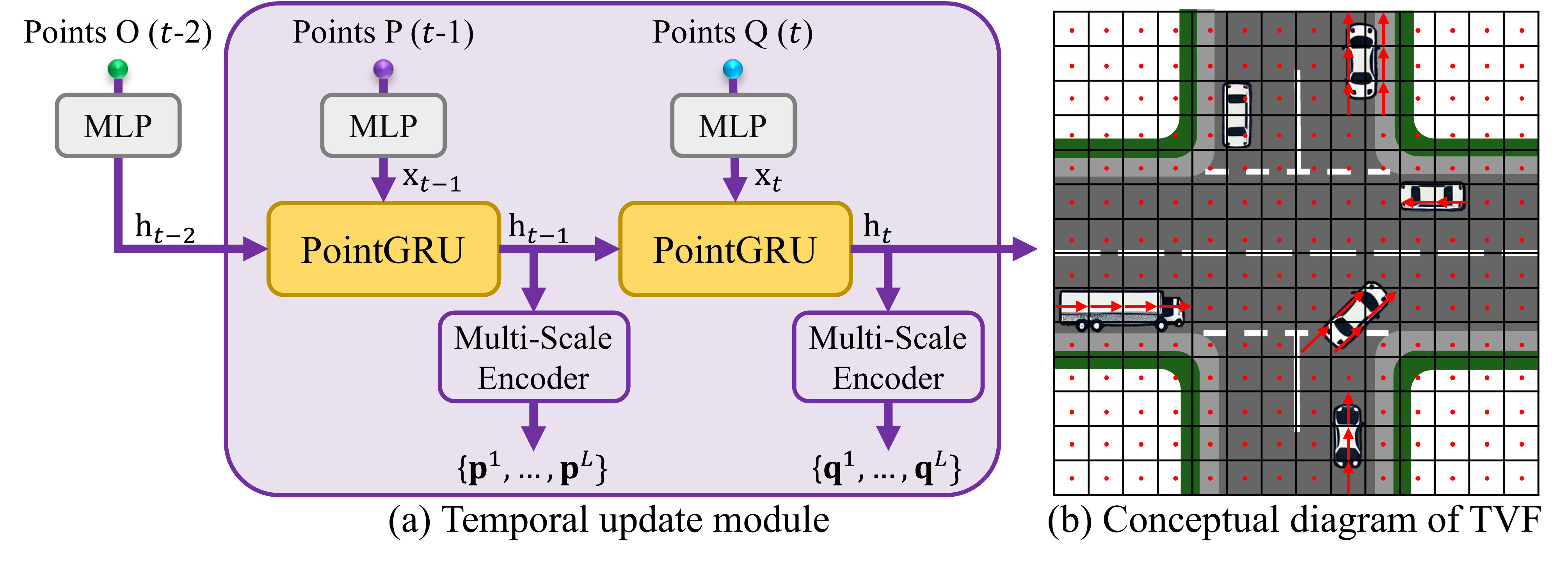}
    \caption{(a) Temporal update module: leverages low-level point dependencies using PointGRU. (b) Conceptual diagram of TVF.} \label{fig:point-GRU}
    \vspace{-0.1cm}
\end{figure}

Recurrent layers can leverage long-range temporal information to enhance radar scene flow. CMFlow uses a GRU to retain flow embeddings across frames but experiences a slight drop in accuracy \cite{CMFlow}. In contrast, we employ PointGRU layers \cite{pointgru} as the temporal module (distinct from TVF encoder's inter-level GRU) to capture dependencies in low-level point features. We initialize the hidden state with the features of point cloud $O$ at time $t-2$ and update it between the current point cloud pairs (see Fig.\ \hyperref[fig:point-GRU]{5a}). During training, we sample sequences of $T$ frames as mini-clips.

    \section{Experiments} \label{sec:exp}

\subsection{Experimental Setup}

\noindent \textbf{Dataset.} We conducted experiments on the VOD dataset \cite{VOD} and a proprietary dataset from Aptiv. Both datasets provide synchronized radar, LiDAR, camera and GPS/IMU odometry data. The VOD dataset contains primarily urban traffic scenes recorded with a low-resolution 4D radar, which captures about 256 radar points per frame and includes 4,662 training samples and 2,724 test samples. The proprietary dataset comprises urban, suburban and highway scenes, merging data from multiple high-resolution radars to obtain about 6K radar points per frame and containing 107,382 training samples and 24,198 test samples.

\noindent \textbf{Metrics.} On the VOD dataset, we adopt the evaluation metrics from CMFlow \cite{CMFlow}:
\begin{enumerate*}
\item \textit{EPE}: mean end-point-error ($L_2$ distance) between the ground truth (GT) and the predicted scene flow.
\item \textit{AccS/AccR}: Strict/Relaxed Accuracy, the percentage of points with EPE $<$ 0.05/0.1 m or a relative error $<$ 5\%/10\%.
\item \textit{RNE}: resolution normalized EPE, to accommodate low-resolution radar. 
\item \textit{MRNE} and \textit{SRNE}: RNE computed separately for moving and static points.
\end{enumerate*}

The proprietary dataset was collected using high-resolution radars. Therefore, we omit RNE. Since moving objects are critical in real-world autonomous driving, we focus on the accuracy of moving points. For moving points, we measure the following:
\begin{enumerate*}
\item \textit{MEPE}: EPE of moving points.
\item \textit{MagE} and \item \textit{DirE}: magnitude and direction error between the GT and prediction.
\item \textit{AccS/AccR} of moving points. 
For static points, we only calculate \item \textit{SEPE}; and we use \item \textit{AvgEPE}, the mean of MEPE and SEPE, as an overall metric.
\end{enumerate*}
Metric details are provided in \cref{appx:metrics}.

\subsection{Implementation Details}

\noindent \textbf{Model details.} The hyperparameters for both datasets are listed in \cref{tab:hyperparams}. Since the VOD dataset has only 256 points per frame, we do not perform downsampling. Our TVF uses a coarse grid to gain a high-level understanding rather than being confined to point-level details. On VOD, we use the Adam optimizer with a learning rate $10^{-3}$, a decay rate of 0.9 per epoch, over 60 epochs. On the proprietary dataset, which is $20\times$ larger, we set the learning rate to $10^{-4}$ with a decay of 0.8 per 30K steps, training for 3 epochs. In the Appendix, we provide details of the OD branch (\ref{appx:od_branch},\ \ref{appx:od_ablation}), runtime analysis (\ref{appx:runtime}) and dual-task training strategy (\ref{appx:joint}).

\begin{table}[tbp]
    \caption{Params for two datasets. $L,N,M, T$: number of levels, points or mini-clips. $\gamma$: downsampling factor. $C, D, D_{\textup{TVF}}$: feature channels of points, flow embeddings, or TVF. $\omega$: number of axial attention blocks. $\mathcal{N}_Q$$,\mathcal{N}_{\textup{TVF}}$: KNN points or TVF cells. TVF grid: shape $[H, W]$ and grid size. ego-info: availability of ego-motion.}
    \label{tab:hyperparams}
    \resizebox{0.48\textwidth}{!}{
    \large
\begin{tabular}{c|c|c|c|c|c|c|c|c|c|c|c|c}
\hline
\hline
Dataset     & $L$ & $N,M$ & $\gamma$ & $C$ & $D$ & $D_{\textup{TVF}}$ & $\omega$ & $\mathcal{N}_Q$ & $\mathcal{N}_{\textup{TVF}}$ & T                & TVF grid      & ego-info \\ \hline
VOD         & 4   & 256   & 1        & 64  & 256 & 128                & 4        & 16              & 9               & 5                & [40,40] 1.28m & Sup.  \\
proprietary & 4   & 6K    & 2        & 64  & 256 & 128                & 4        & 8               & 9               & 12               & [70,40] 2.0m  & Input      \\ \hline
\hline
\end{tabular}
}
\end{table}

\begin{table}[htbp]
    \caption{Model variants and supervision signals.}
    \vspace{-0.15cm}
    \label{tab:model_variants}
    \resizebox{0.48\textwidth}{!}{
    \large
\begin{tabular}{lccc|cc}
\hline
\hline
\multicolumn{1}{l|}{\multirow{2}{*}{TARS}} & \multicolumn{3}{c|}{EM (ego-motion $\Omega$) availability}                & \multicolumn{2}{c}{Supervision}                                                                                                   \\ \cline{2-6} 
\multicolumn{1}{c|}{}                       & train\hspace{-2pt}  & \multicolumn{1}{c|}{\hspace{-2pt}test}   & Note                               & \multicolumn{1}{c|}{Label}       & Note                                                                                           \\ \hline
\multicolumn{1}{l|}{ego}               & \ding{51}\hspace{-2pt} & \multicolumn{1}{c|}{\hspace{-2pt}\ding{55}} & as GT to train an EM head  & \multicolumn{1}{c|}{Cross}       & w/ all losses                                                                     \\
\multicolumn{1}{l|}{superego}          & \ding{51}\hspace{-2pt} & \multicolumn{1}{c|}{\hspace{-2pt}\ding{51}} & as input to compensate EM & \multicolumn{1}{c|}{Cross$^{+}$} & w/o $\mathcal{L}_{\textup{seg}}$, $\mathcal{L}_{\textup{ego}}, \mathcal{L}_{\textup{opt}}$ \\
\multicolumn{1}{l|}{no-ego}            & \ding{55}\hspace{-2pt} & \multicolumn{1}{c|}{\hspace{-2pt}\ding{55}} & no EM operation  & \multicolumn{1}{c|}{/}           & w/o $\mathcal{L}_{\textup{seg}}$, $\mathcal{L}_{\textup{ego}}, \mathcal{L}_{\textup{opt}}$ \\ \hline
\multicolumn{4}{c|}{FlowStep3D \cite{FlowStep3D}, RaFlow \cite{RaFlow}, etc.}                                                                                & \multicolumn{1}{c|}{Self}        & w/ only $\mathcal{L}_{\textup{sc}}$, $\mathcal{L}_{\textup{ss}}, \mathcal{L}_{\textup{rd}}$ \\ \hline
\hline
\end{tabular}
}
\vspace{-0.15cm}
\end{table}

On the proprietary dataset, we simulate real-world autonomous driving by providing \textbf{all} models with ego-motion $\Omega \in \mathbb{R}^{4\times4}$ from the GPS/IMU sensor as known input. We apply ego-motion compensation to align $P$ and $Q$ into the same coordinate system.  In this case, the GT for static points is the zero vector. On the VOD dataset, we test our model under two setups:
\begin{enumerate*}[label=(\roman*)]
\item \textbf{TARS-ego:} following CMFlow \cite{CMFlow}, using the ego-motion transformation to train an additional ego-motion head for a fair comparison;
\item \textbf{TARS-superego:} using ego-motion as known input and applying compensation, same as on the proprietary dataset.
\end{enumerate*} For details of the setups, please see \cref{appx:ego_head} and \ref{appx:superego}.

\begin{table*}[tbp]
\centering
    \caption{Scene flow evaluation on the VOD dataset. Mean metric values across the test set are reported. ``Sup.'' indicates the supervision signal, Self: training with only self-supervised losses \cite{RaFlow}, Cross: with additional cross-modal losses \cite{CMFlow}, Cross$^+$: setup for TARS-superego, without $\mathcal{L}_{\textup{seg}}$, $\mathcal{L}_{\textup{ego}}$, and $\mathcal{L}_{\textup{opt}}$ while using ego-motion $\Omega$ as known input for ego-motion compensation, same as on the proprietary dataset.}
    \vspace{-0.15cm}
    \label{tab:VOD}
    \resizebox{0.85\textwidth}{!}{
    \large
\begin{tabular}{lc|cccccc|ccc|ccc}
\hline
\hline
                                          & \multicolumn{1}{l|}{} & \multicolumn{6}{c|}{Overall}                                                                                                                                                                                                                                        & \multicolumn{3}{c|}{Moving}                                          & \multicolumn{3}{c}{Static}                                           \\ \hline
\multicolumn{1}{c|}{Method}               & Sup.                  &  & \quad EPE [m]$\downarrow$ \quad                              & \quad AccS [\%]$\uparrow$ \quad                            & \quad AccR [\%]$\uparrow$ \quad                            & \quad \enspace RNE [m]$\downarrow$ \enspace \quad &  &  & \quad MRNE [m]$\downarrow$ \quad                             &  &  & \quad SRNE [m]$\downarrow$ \quad                             &  \\ \hline
\multicolumn{1}{l|}{PointPWC-Net \cite{PointPWC}}         & Self                  &  & 0.422                                                          & 2.6                                                          & 11.3                                                         & 0.169                                                          &  &  & 0.154                                                          &  &  & 0.170                                                          &  \\
\multicolumn{1}{l|}{SLIM \cite{SLIM}}               & Self                  &  & 0.323                                                          & 5.0                                                         & 17.0                                                         & 0.130                                                          &  &  & 0.151                                                          &  &  & 0.126                                                          &  \\
\multicolumn{1}{l|}{FlowStep3D \cite{FlowStep3D}}           & Self                  &  & 0.292                                                          & 3.4                                                          & 16.1                                                         & 0.117                                                          &  &  & 0.130                                                          &  &  & 0.115                                                          &  \\
\multicolumn{1}{l|}{Flow4D-2frame \cite{Flow4D}}               & Cross                  &  & 0.255                                                          & 10.0                                                         & 26.2                                                         & 0.103                                                          &  &  & 0.125                                                          &  &  & 0.098                                                          &  \\
\multicolumn{1}{l|}{RaFlow \cite{RaFlow}}               & Self                  &  & 0.226                                                          & 19.0                                                         & 39.0                                                         & 0.090                                                          &  &  & 0.114                                                          &  &  & 0.087                                                          &  \\
\multicolumn{1}{l|}{DeFlow \cite{DeFlow}}               & Cross                  &  & 0.217                                                          & 11.8                                                         & 31.6                                                         & 0.087                                                          &  &  & 0.098                                                          &  &  & 0.085                                                          &  \\
\multicolumn{1}{l|}{CMFlow \cite{CMFlow}}               & Cross                 &  & 0.130                                                          & 22.8                                                         & 53.9                                                         & 0.052                                                          &  &  & 0.072                                                          &  &  & 0.049                                                          &  \\
\multicolumn{1}{l|}{TARS-ego (ours)}      & Cross                 &  & \makebox[5ex][l]{\textbf{0.092} \textcolor{mygreen}{(-0.038)}} & \makebox[4ex][l]{\textbf{39.0} \textcolor{mygreen}{(+16.2)}} & \makebox[4ex][l]{\textbf{69.1} \textcolor{mygreen}{(+15.2)}} & \makebox[5ex][l]{\textbf{0.037} \textcolor{mygreen}{(-0.015)}} &  &  & \makebox[5ex][l]{\textbf{0.061} \textcolor{mygreen}{(-0.011)}} &  &  & \makebox[5ex][l]{\textbf{0.034} \textcolor{mygreen}{(-0.015)}} &  \\ \hline
\multicolumn{1}{l|}{TARS-superego (ours)} & Cross$^{+}$           &  & \textbf{0.048}                                                 & \textbf{76.6}                                                & \textbf{86.4}                                                & \textbf{0.019}                                                 &  &  & \textbf{0.057}                                                 &  &  & \textbf{0.014}                                                 &  \\ \hline
\hline
\end{tabular}
}
\end{table*}

\begin{table*}[htbp]
\centering
    \caption{Scene flow evaluation on the proprietary dataset. Mean metric values across the test set are reported. Ego-motion compensation and Cross$^{+}$ setup are applied to \textbf{all} models, making them ``superego''. PointGRU is applied in all models except for the one in the first row.}
    \vspace{-0.15cm}
    \label{tab:proprietary}
    \resizebox{0.9\textwidth}{!}{
    \large
\begin{tabular}{lc|lcccccl|lcl|lcl}
\hline
\hline
                                                  &              &  & \multicolumn{5}{c}{Moving}                                                                                                                                                                                                                                                                                           &  &  & Overall                                                      &  &  & Static                   &  \\ \hline
\multicolumn{1}{c|}{Method}                        & PointGRU    &  & \quad MEPE [m]$\downarrow$ \quad                                  & \quad MagE [m]$\downarrow$ \quad                             & \quad DirE [rad]$\downarrow$ \quad                           & \quad AccS [\%]$\uparrow$ \quad                            & \quad AccR [\%]$\uparrow$ \quad                                   &  &  & AvgEPE [m]$\downarrow$                                       &  &  & \quad SEPE [m]$\downarrow$     &  \\ \hline
\multicolumn{1}{l|}{PointPWC-Net \cite{PointPWC}} & \ding{55}      &  & 0.453                                                       & 0.363                                                        & 1.218                                                        & 44.2                                                       & 52.2                                                        &  &  & 0.244                                                        &  &  & 0.036                    &  \\
\multicolumn{1}{l|}{PointPWC-Net \cite{PointPWC}}                 & \ding{51} &  & 0.213                                                       & 0.178                                                        & 0.762                                                        & 49.0                                                       & 60.5                                                        &  &  & 0.124                                                        &  &  & \textbf{0.035}           &  \\
\multicolumn{1}{l|}{HALFlow \cite{HALFlow}}       & \ding{51} &  & 0.170                                                       & 0.135                                                        & 0.721                                                        & 50.9                                                       & 63.8                                                        &  &  & 0.104                                                        &  &  & 0.038                    &  \\
\multicolumn{1}{l|}{TARS (ours)}                  & \ding{51} &  & \makebox[5ex][l]{\textbf{0.069} \textcolor{mygreen}{(-0.101)}} & \makebox[6ex][l]{ \textbf{0.059} \textcolor{mygreen}{(-0.076)}} & \makebox[6ex][l]{ \textbf{0.599} \textcolor{mygreen}{(-0.122)}} & \makebox[4ex][l]{\textbf{69.8} \textcolor{mygreen}{(+18.9)}} & \makebox[5ex][l]{ \textbf{86.8} \textcolor{mygreen}{(+23.0)}} &  &  & \makebox[6ex][l]{ \textbf{0.054} \textcolor{mygreen}{(-0.05)}} &  &  & \makebox[6ex][l]{ 0.038} &  \\ \hline
\hline
\end{tabular}
}
\vspace{-0.2cm}
\end{table*}

\noindent \textbf{Weakly-supervised training.} Since annotating scene flow is extremely difficult, we adopt the self-supervised losses in \cite{RaFlow} and cross-modal losses in \cite{CMFlow}, and introduce an additional background loss for static points. Detailed loss functions are given in \cref{appx:loss}. On both datasets, we apply the following losses: 
1.\ soft Chamfer loss $\mathcal{L}_{\textup{sc}}$: aligns $P_{\textup{warp}}$ and $Q$ by minimizing nearest‐point distances, handling outliers via probabilistic matching; 2.\ spatial smoothness loss $\mathcal{L}_{\textup{ss}}$: enforces neighboring points to have similar flow vectors, weighted by distance to ensure spatial smoothness; 3.\ radial displacement loss $\mathcal{L}_{\textup{rd}}$: constrains the radial projection of predicted flow vectors using radar’s RRV measurements; 4.\ foreground loss $\mathcal{L}_{\textup{fg}}$: derives the pseudo scene flow GT from a LiDAR multi-object tracking model, applied to the predicted flow $F^{l}_{\textup{fg}}$ of foreground moving points at each level; 5.\ additionally, we employ a background loss $\mathcal{L}_{\textup{bg}}$: using the ego-motion transformation as pseudo GT $\hat{F}^{l}_{\textup{bg}}$ for static points. The overall loss $\mathcal{L}_{\textup{all}}$ is formulated as:
\begin{gather}
     \mathcal{L}_{\textup{all}} =\mathcal{L}_{\textup{sc}} + \mathcal{L}_{\textup{ss}} + \mathcal{L}_{\textup{rd}} + \sum\nolimits_{l=1}^{L}(\mathcal{L}^l_{\textup{fg}} + \lambda_{\textup{bg}}\mathcal{L}^l_{\textup{bg}}),
\end{gather}
where $\mathcal{L}^l_{\textup{fg}}$ and $\mathcal{L}^l_{\textup{bg}}$ are computed for each level, $\lambda_{\textup{bg}}=0.5$.

When training TARS-ego on the VOD dataset, we incorporate cross-modal losses \cite{CMFlow}: motion segmentation loss $\mathcal{L}_{\textup{seg}}$: uses pseudo segmentation GT (derived from odometer and RRV measurements) to train a motion-segmentation head; ego-motion loss $\mathcal{L}_{\textup{ego}}$: uses GT ego-motion to train an ego-motion head; and optical flow loss $\mathcal{L}_{\textup{opt}}$: projects the scene flow onto the image plane and is trained with pseudo optical flow labels. \cref{tab:model_variants} summarizes the variants of TARS and their supervision signals.

\subsection{Comparison with State of the Art}

\noindent \textbf{Experiments on the VOD dataset.} We compare our model TARS with SOTA scene flow methods on the VOD dataset (see \cref{tab:VOD}). TARS clearly outperforms the previous SOTA model CMFlow \cite{CMFlow} across all evaluation metrics in both setups. Under the same setup as CMFlow, our TARS-ego reduces the overall EPE from $0.13\text{m}$ to $0.092\text{m}$, marking a $0.038\text{m}$ reduction and achieving a new milestone by bringing EPE below the AccR threshold of $0.1\text{m}$. Moreover, TARS-ego improves AccS and AccR, two accuracy metrics computed based on EPE, by $16.2\%$ and $15.2\%$, respectively. RNE is computed as $RNE= \frac{EPE}{r_R / r_L}$, where $\frac{r_R}{r_L}$ is the ratio of radar to LiDAR resolution (on average 2.5). Therefore, reductions in RNE metrics are numerically smaller. Nevertheless, TARS-ego shows substantial improvements in all three RNE-related metrics. We also compare TARS-ego with latest LiDAR models \cite{Flow4D, DeFlow}. These models perform worse because their fully-voxel representation, designed for large-scale LiDAR point clouds, is unsuitable for radar scene flow. Ego-motion from the odometer is a simple $\mathbb{R}^{4\times4}$ matrix yet effective booster. Assuming real-world autonomous driving with an available odometer, TARS-superego applies ego-motion compensation and reduces EPE down to $0.048\text{m}$, MRNE to $0.057\text{m}$, and boosts AccS and AccR to $76.6\%$ and $86.4\%$, respectively.

\begin{figure*}[t] \centering

    \rotatebox{90}{\scriptsize ~~~~~~~~~~~TARS (ours) ~~~~~~~~~~~~~~~~~~~~~ HALFlow}
    \includegraphics[width=0.97\textwidth]{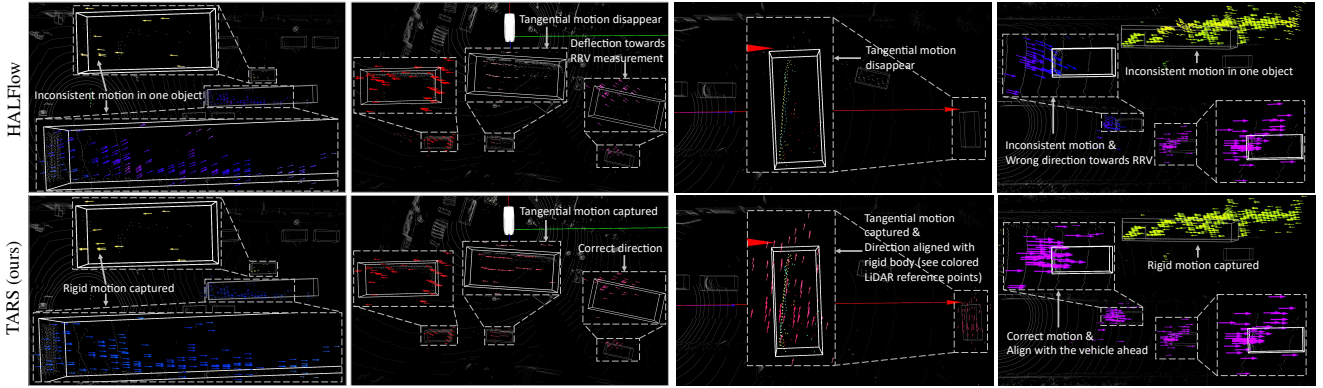}

    % \vspace{-0.1cm}
    \caption{Qualitative results on the proprietary dataset, compared with HALFlow \cite{HALFlow}. LiDAR point clouds serve as reference. Arrows indicate the predicted scene flow. After ego-motion compensation, static points are expected to yield zero vectors if predicted correctly. By perceiving traffic-level motion cues, TARS effectively captures the rigid motion in Euclidean space, as well as tangential movements.} \label{fig:quality}
\vspace{-0.3cm}
\end{figure*}

\noindent \textbf{Experiments on the proprietary dataset.} The proprietary dataset includes more complex and high-speed scenarios such as highways and suburban scenes, which makes radar scene flow estimation particularly challenging. Therefore, to simulate real-world autonomous driving, we apply ego-motion compensation to \textbf{all} models and focus on moving points during evaluation (see \cref{tab:proprietary}). Furthermore, we integrate our temporal update module (\cref{sec:3.5}) into other models for a fair comparison. Our model TARS outperforms the previous SOTA model HALFlow \cite{HALFlow} by a large margin, reducing MEPE from $0.17\text{m}$ to $0.069\text{m}$ and improving AccS and AccR by $18.9\%$ and $23\%$, respectively. Qualitative results in \cref{fig:quality} show that TARS effectively captures the rigid motion of radar points on the same object and mitigates the tangential motion challenge mentioned in \cref{sec:intro}.

\subsection{Ablation Studies}

\begin{table}[tbp]
    \caption{Ablation study on the proprietary dataset. Point Level: point-level motion understanding. OD Featmap: using OD feature map. Coarse Grid: using coarse-grid TVF. Glob Attn: global attention applied. The chosen setup of TARS is highlighted in blue.}
    \label{tab:ablation_prop}
    \resizebox{0.48\textwidth}{!}{
    \large
\begin{tabular}{c|cccc|ccc|c|c}
\hline
\hline
\multicolumn{1}{l|}{\multirow{2}{*}{No.}} & Point                         & OD                            & Coarse                        & 	Glob                         & \multicolumn{3}{c|}{Moving}                        & Overall            & Static           \\ \cline{6-10} 
\multicolumn{1}{l|}{}                     & Level                         & Featmap                       & Grid                          & Attn                          & MEPE$\downarrow$ & AccS$\uparrow$ & AccR$\uparrow$ & AvgEPE$\downarrow$ & SEPE$\downarrow$ \\ \hline
1                                         & \ding{51}                       &                                &                               &                               & 0.178            & 47.9           & 61.6           & 0.106              & 0.033            \\
2                                         & \ding{51}                       & \ding{55}                       & \ding{51}                       & \ding{51}                       & 0.144            & 45.0           & 63.3           & 0.093              & 0.041            \\
3                                         & \ding{51}                       & \ding{51}                       & \ding{55}                       & \ding{51}                       & 0.104            & 51.4           & 69.9           & 0.076              & 0.049            \\
4                                         & \ding{51}                       & \ding{51}                       & \ding{51}                       & \ding{55}                       & 0.074            & 65.6           & 84.2           & \textbf{0.053}              & \textbf{0.031}   \\
5                                         & \textcolor{myblue}{\ding{51}} & \textcolor{myblue}{\ding{51}} & \textcolor{myblue}{\ding{51}} & \textcolor{myblue}{\ding{51}} & \textbf{0.069}   & \textbf{69.8}  & \textbf{86.8}  & 0.054     & 0.038            \\ \hline
6                                         & \multicolumn{4}{c|}{Decoder $\mathcal{N}_{\textup{TVF}}=5\enspace$}                                                           & 0.071            & 67.2           & 86.5           & 0.064              & 0.057            \\
7                                         & \multicolumn{4}{c|}{Decoder $\textcolor{myblue}{\mathcal{N}_{\textup{TVF}}=9}\enspace$}                                       & \textbf{0.069}   & \textbf{69.8}  & \textbf{86.8}  & \textbf{0.054}     & \textbf{0.038}   \\
8                                         & \multicolumn{4}{c|}{Decoder $\mathcal{N}_{\textup{TVF}}=13$}                                                                  & 0.077            & 59.9           & 84.0           & 0.060              & 0.043            \\ \hline
\hline
\end{tabular}
}
\vspace{-0.2cm}
\end{table}

\noindent \textbf{Importance of key components.} We demonstrate the effectiveness of the key components of TARS on the proprietary dataset. %as shown in \cref{tab:ablation_prop}. 
Ablation study \textit{No.}\ $i$ refers to the $i$-th row of \cref{tab:ablation_prop}.  \textit{No.\,1}: We completely remove the traffic level, causing the model to revert to inferring motion cues solely at the point level, resulting in MEPE of $0.178\text{m}$, close to HALFlow (\cref{tab:proprietary}). \textit{No.\,2}: We activate the traffic level but without the OD feature map, meaning no scene update is performed in the TVF encoder. This reduces our model to simply combining point- and voxel-based motion features, leading to unsatisfactory AccR of $63.3\%$. \textit{No.\,3}: We enable the OD feature map and scene update, but change the coarse grid of TVF to a fine grid, from $2\mathrm{m} \times 2\mathrm{m}$ to $1\mathrm{m} \times 1\mathrm{m}$ with shape $[140, 80]$. In this case, our traffic-level understanding falls into point-level details, resulting in a reduction of MEPE by $0.04\text{m}$, yet the improvement in accuracy is limited (AccR $69.9\%$). \textit{No.\,4}: We apply a coarse TVF to achieve high-level understanding, which further reduces MEPE by $0.03\text{m}$ and boosts the accuracy; but we replace the global attention in the TVF encoder with local convolutions. This makes the TVF focus on local areas, achieving the lowest SEPE of $0.031\text{m}$, while limiting its ability to capture global traffic information (e.g., motion of vehicles in the same lane). \textit{No.\,5}: We construct a holistic traffic model using global attention. Compared to No.\,4, this improves AccS by $4.2\%$ and AccR by $2.6\%$. Although SEPE increased, all models maintain a low SEPE (below $0.05\text{m}$) due to ego-motion compensation, and we prioritize dynamic objects in autonomous driving. \textit{No.\,6-8} show the effect of $\mathcal{N}_{\text{TVF}}$: the number of TVF cells that a point query attends to, when capturing motion context. In the cross-attention, considering more spatial context $\mathcal{N}_{\text{TVF}} = 9$ (surrounding neighbors) improves AccS by $2.6\%$ and reduces SEPE by $0.019\text{m}$, compared to $\mathcal{N}_{\text{TVF}} = 5$ (only direct neighbors). However, expanding to $\mathcal{N}_{\text{TVF}} = 13$ (second-order neighbors) increases SEPE and significantly reduces AccS due to the inclusion of irrelevant parts in the attention. Moreover, on PointPWC-Net in \cref{tab:proprietary}, we demonstrate the importance of applying PointGRU to utilize temporal information. 

\begin{table}[tbp]
    \caption{Ablation study on the VOD dataset. TARS-no-ego: without ego-motion head and supervision signals $\mathcal{L}_{\{ \textup{seg, ego, opt} \}}$. The chosen setup of TARS-ego is highlighted in blue.}
    \label{tab:ablation_vod}
    \resizebox{0.48\textwidth}{!}{
    \large
\begin{tabular}{c|c|cc|cccc|c|c}
\hline
\hline
\multirow{2}{*}{No.} & \multirow{2}{*}{TARS} & $\{ \mathcal{L}_{\textup{seg}},$                            & $\mathcal{L}_{\textup{bg}} $ & \multicolumn{4}{c|}{Overall}                                        & Moving           & Static           \\ \cline{5-10} 
                     &                       & $\mathcal{L}_{\textup{ego}}, \mathcal{L}_{\textup{opt}} \}$ & or $\lambda_{\textup{bg}}$     & EPE$\downarrow$ & AccS$\uparrow$ & AccR$\uparrow$ & RNE$\downarrow$ & MRNE$\downarrow$ & SRNE$\downarrow$ \\ \hline
\multirow{4}{*}{1}   & no-ego                & \ding{55}                                                     & 0.25                         & 0.124           & 23.6           & 54.7           & 0.050           & 0.066            & 0.048            \\
                     & no-ego                & \ding{55}                                                     & \textcolor{myblue}{0.50}     & 0.111           & 28.5           & 59.3           & 0.045           & \textbf{0.065}   & 0.043            \\
                     & no-ego                & \ding{55}                                                     & 0.75                         & 0.103           & 32.5           & 63.2           & 0.042           & 0.066            & 0.039            \\
                     & no-ego                & \ding{55}                                                     & 1.00                         & \textbf{0.098}  & \textbf{34.3}  & \textbf{65.8}  & \textbf{0.040}  & 0.067            & \textbf{0.036}   \\ \hline
\multirow{2}{*}{2}   & ego                   & \ding{51}                                                     & \ding{55}                      & 0.107           & 32.5           & 62.4           & 0.043           & 0.062            & 0.040            \\
                     & \textcolor{myblue}{ego}                   & \textcolor{myblue}{\ding{51}}                               & \textcolor{myblue}{0.50}     & \textbf{0.092}  & \textbf{39.0}  & \textbf{69.1}  & \textbf{0.037}  & \textbf{0.061}   & \textbf{0.034}   \\ \hline
\multirow{3}{*}{3}   & ego                   & \multicolumn{2}{l|}{Decoder $\mathcal{N}_{\textup{TVF}}=5$}                                & 0.094           & 38.8           & 68.5           & 0.038           & 0.062            & 0.034            \\
                     & ego                   & \multicolumn{2}{l|}{Decoder $\textcolor{myblue}{\mathcal{N}_{\textup{TVF}}=9}$}            & \textbf{0.092}  & \textbf{39.0}  & \textbf{69.1}  & \textbf{0.037}  & \textbf{0.061}   & 0.034            \\
                     & ego                   & \multicolumn{2}{l|}{Decoder $\mathcal{N}_{\textup{TVF}}=13$}                               & 0.093           & 38.1           & 68.7           & 0.037           & 0.062            & 0.034            \\ \hline
\hline
\end{tabular}
}
\vspace{-0.2cm}
\end{table}

\noindent \textbf{Effect of losses.} We evaluate the impact of losses on the VOD dataset (\cref{tab:ablation_vod}). Group \textit{No.\,1}: we test the impact of the proposed background loss $\mathcal{L}_{\text{bg}}$ using a TARS-no-ego model, which excludes the ego-motion head of TARS-ego as well as three losses $\mathcal{L}_{\text{seg}}$, $\mathcal{L}_{\text{ego}}$, and $\mathcal{L}_{\text{opt}}$. The experiments show that setting the background weight $\lambda_{\text{bg}}=0.5$ results in the lowest MRNE of moving points. Further increasing $\lambda_{\text{bg}}$ could inflate AccS\&AccR as they reflect overall accuracy on the VOD dataset, where static points dominate the scene. However, it undermines the MRNE. We advocate emphasizing dynamic objects in radar scene flow, which is why we set $\lambda_{\text{bg}} = 0.5$. Group \textit{No.\,2}: we include the ego-motion head but initially without $\mathcal{L}_{\text{bg}}$, which yields AccR of $62.4\%$. After enabling $\mathcal{L}_{\text{bg}}$, both moving and static points get improved. Group \textit{No.\,3}: we test the impact of $\mathcal{N}_{\text{TVF}}$. However, it did not yield significant differences among the three setups, because each frame in the VOD dataset contains only 256 points, resulting in a highly sparse TVF.

    \section{Conclusion} \label{sec:conclusion}
We introduced TARS, a traffic-aware radar scene flow model. Leveraging traffic information from an object detector, TARS employs the traffic vector field encoder to build a holistic traffic-level scene understanding and uses the TVF decoder to perceive motion rigidity in the traffic context. Quantitative results show that TARS significantly outperforms SOTA on both the proprietary dataset and VOD dataset. Ablation studies highlight the effectiveness of key components in our design, such as incorporating OD feature maps, using a coarse-grid TVF, and applying global attention. Qualitative results demonstrate TARS's ability to capture rigid motion and tangential movements.

\clearpage

    \section*{Acknowledgements}
    J.W.\ and M.R.\ acknowledge support by the German Federal Ministry of Education and Research within the junior research group project “UnrEAL” (grant no.\ 01IS22069).
    
    \small
    \putbib    % 输出正文引用
\end{bibunit}

\begin{bibunit}
    \clearpage
\setcounter{page}{1}
\maketitlesupplementary
\appendix
\setcounter{table}{0}
\renewcommand{\thetable}{\Alph{table}}

\section{Addntional Implementation Details}

\subsection{Further Model Details} \label{appx:details}

In this section we provide more details about TARS.

When computing the point-level flow embeddings in \cref{sec:3.1.1}, we use an MLP for positional encoding but omit it in those equations for simplicity. It is also applied in cross-attention in the TVF decoder (\cref{sec:3.4}). 

In the scene update stage in \cref{sec:3.3}, we firstly apply CNN layers to the feature map $\chi_{\textup{od}}^l \in \mathbb{R}^{H_l \times W_l \times D_l}$ to adapt the object features. We then apply pooling layers, because feature maps from the OD pyramid have varying resolutions ($H_l, W_l$) at each level, and we match them with the pre-defined shape of the coarse TVF.

In the flow painting stage in \cref{sec:3.3}, we fuse the scene $\textbf{X}_\textup{traffic}^l$ and motion feature $\textbf{X}_\textup{motion}^l$ in a spatial attention style \cite{SpatialAtt}, specifically:
\begin{flalign} \label{eq:fusion}
     &  \textbf{X}_\textup{fusion}^l = \textbf{w}^l \odot \textbf{X}_\textup{traffic}^l + (1- \textbf{w}^l) \odot \textbf{X}_\textup{motion}^l, \quad \text{with} \\
     & \textbf{w}^l = \sigma( \textup{Concat}( \textbf{W}_{1} \ast \textbf{X}_\textup{traffic}^l, \textbf{W}_{2} \ast \textbf{X}_\textup{motion}^l)),
\end{flalign}
where $ \textbf{W}_1,  \textbf{W}_2$ are CNN kernels that generate pixel-wise attention scores, $\textbf{w}^l$ is the pixel-wise attention weights.

In the flow painting stage in \cref{sec:3.3}, we use axial attention to provide the global receptive field, specifically:
\begin{flalign}
    & \textup{TVF}_\tau^l = \textup{TVF}_\tau^l\textup{-H} + \textup{TVF}_\tau^l\textup{-W},  \quad \text{with} \\
     &  \textup{TVF}_\tau^l\textup{-H} = \textup{Self-A}(\textup{Col}(\textup{TVF}_{\tau-1}^l)), \\ 
     & \textup{TVF}_\tau^l\textup{-W}=\textup{Self-A}(\textup{Row}(\textup{TVF}_\tau^l\textup{-H})), \\
     & \textup{Self-A}(\cdot) = \textup{Attention}(\cdot, \cdot, \cdot),
\end{flalign}
where $\tau=\{1,...,\omega\}$, $\textup{Col}(\cdot)$ collects $\textup{W}$ column vectors and $\textup{Row}(\cdot)$ collects $\textup{H}$ row vectors from the 2D TVF.

On the proprietary dataset, we provide \textbf{all} models with ego-motion $\Omega \in \mathbb{R}^{4\times4}$ as known input. We apply ego-motion compensation before the PointGRU blocks (\cref{sec:3.5}), aligning $O$ and $P$, as well as $P$ and $Q$ into the same coordinate system.

\subsection{Ego-Motion Head Details} \label{appx:ego_head}
On the VOD dataset, our TARS-ego adopts the same approach as RaFlow and CMFlow \cite{RaFlow, CMFlow} to train an ego-motion head. It predicts the ego transformation, which is treated as static flow and assigned to all static points. This assignment requires our model to distinguish between moving and static points, which is achieved by a motion-segmentation head.

\noindent \textbf{Motion-segmentation head.} 
The motion-segmentation head takes the final flow embeddings from the $L$-th level of TARS-ego as input and outputs a probability map $\textbf{S}$ indicating for each point whether it is moving or static. We use a simple three-layer MLP followed by a sigmoid function to generate the probability map. A binary segmentation mask can be generated from $\textbf{S}$ using a threshold of 0.5.

We apply the same training strategy in previous works \cite{RaFlow, CMFlow}. During training, we use ground truth (GT) segmentation pseudo-labels as the segmentation mask, in order to provide stable segmentation results for training the ego-motion head. During testing, we use the model output from the motion-segmentation head.

\noindent \textbf{Ego-motion head.} Since $P_{\textup{warp}}$ is obtained by warping $P$ using the scene flow output $F^L$ from the last $L$-th level, they can be used as natural correspondences to infer the ego motion. In the ego-motion head, we take $P$, $P_{\textup{warp}}$, and $\textbf{S}$ as input. Then we perform the differentiable weighted Kabsch algorithm \cite{Kabsh}, which infers rigid transformation between $P$ and $P_{\textup{warp}}$, with $(1-\textbf{S})$ as the point weight to neglect moving points. Finally, we use the binary motion segmentation mask generated from $\textbf{S}$ to identify static points and assign the inferred ego-motion to static points as their final scene flow $F_{\textup{bg}}$. Since the ego-motion is inferred between $P$ and $P_{\textup{warp}}$, a more accurate $F^L$ leads to improved ego-motion predictions and, ultimately, more accurate static flow $F_{\textup{bg}}$. Results of the ego-motion estimation are shown in \cref{tab:ego_result}.

\begin{table}[htbp]
\centering
    \caption{Evaluation of the ego-motion head on VOD. RTE: relative translation error. RAE: relative angular error.}
    \label{tab:ego_result}
    \resizebox{0.27\textwidth}{!}{
    \large
\begin{tabular}{l|cc}
\hline
\hline
Model                                 & RTE {[}m{]}    & RAE {[}\textdegree{]} \\ \hline
CMFlow \cite{CMFlow} & 0.083          & 0.140  \\ \cline{1-3}
TARS-ego (ours)                              & \textbf{0.059} & \textbf{0.098}                       \\ \hline
\hline
\end{tabular}
}
\end{table}

\subsection{TARS-ego and TARS-superego} \label{appx:superego}
On the VOD dataset, we evaluate our model using two setups that apply the ego-motion information differently: TARS-ego and TARS-superego, results shown in \cref{tab:VOD} of the main manuscript. TARS-ego uses GT ego-motion to train a ego-motion head and motion-segmentation head to refine the static flow $F_{\textup{bg}}$; while TARS-superego uses ego-motion as input and applies ego-motion compensation, expecting the static flow $F_{\textup{bg}}$ to be zero vectors (without those additional heads). On the proprietary dataset, since ego-motion compensation is applied to \textbf{all} models, we do not specifically label our model as TARS-superego. In this section, we discuss the reason of leveraging ego-motion.

On one hand, ego-motion information from an odometer is simple, easy to obtain, yet highly effective. Utilizing this information enhances scene flow performance in both ways: as supervision signal for an ego-motion head, or as known input for ego-motion compensation. Nevertheless, the latter aligns more closely with real-world autonomous driving scenarios, where a GPS/IMU sensor is often available. It helps address the challenges of radar scene flow.

On the other hand, performing ego-motion compensation maximizes the potential of our model TARS. By compensating point cloud $P$ into the coordinate system of point cloud $Q$, the points in $P$ and the OD feature map from $Q$ share a common coordinate system, starting from the lowest level of TARS. This eliminates the need for the lower levels of TARS to firstly warp the points in $P$ closer to the corresponding object features, ensuring alignment within the TVF throughout the process.

\begin{algorithm}[htbp]
\caption{TARS-ego Forward Pass}
\label{alg:main}
\begin{algorithmic}[1]
\Require Consecutive point clouds $P, Q$, previous point cloud $O$ and its low-level hidden state $h_{t-2}$
\Ensure Scene flow $F^L$, motion segmentation mask $\mathbf{S}$, predicted ego-motion $\hat{\Omega}$, hidden state $h_{t-1}$
\State \Comment{Collect multi-level OD features}
\State $\{\chi^l_{\textup{od}}\}_{l=1}^L \gets \text{ODBranch}(Q)$

\State \Comment{1. Point feature extraction}
\State $(P,Q,\mathbf{p},\mathbf{q}) \gets \text{MLP}(P,Q)$

\State \Comment{2. Low-level temporal module}
\State $(P,Q,\mathbf{p},\mathbf{q},h_{t-1}) \gets \text{PointGRU}(P,Q,\mathbf{p},\mathbf{q},O,h_{t-2})$

\State \Comment{3. Point patch feature extraction}
\State \Comment{when $l=L$: full point set}
\State $\{P^l,Q^l,\mathbf{p}^l,\mathbf{q}^l\}_{l=1}^L \gets \text{MultiScaleEncoder}(P, Q,\mathbf{p},\mathbf{q})$

\State \Comment{4. Flow \& embeddings at $l=1$}
\State $P^{1}_{\textup{warp}} \gets P^1 + \mathbf{0}$
\State $\mathbf{e}_{\textup{point}}^{1}\gets \textup{PointLevelEmb}(P^{1}_{\textup{warp}},Q^1,\mathbf{p}^1,\mathbf{q}^1)$

\State $(F^1,\mathbf{e}^1)\gets \textup{FlowHead}(\mathbf{e}_{\textup{point}}^{1},\varnothing,\varnothing)$

\State \Comment{5. Coarse-to-fine prediction}
\State $\textup{TVF}^{1}\gets \varnothing$ 
\For{$l=2$ up to $L$}
  \State $\textup{TVF}^{l} \gets \textup{TVF\_Encoder}(P^{l-1},\mathbf{p}^{l-1}, F^{l-1},$
  \State \quad\quad\quad\quad\quad\quad\quad\quad\quad\quad\quad $\mathbf{e}^{l-1},\chi^{l}_{\textup{od}},\textup{TVF}^{l-1})$

  \State $P^{l}_{\textup{warp}} \gets P^l + \textup{Interp}(F^{l-1})$

  \State $\mathbf{e}_{\textup{point}}^{l}\gets \textup{PointLevelEmb}(P^{l}_{\textup{warp}},Q^l,\mathbf{p}^l,\mathbf{q}^l)$

  \State $\hat{\textbf{p}}^l = \textup{Concat}(\textup{Interp}(\textbf{e}^{l-1}), \textbf{p}^l)$
  \State $\mathbf{e}_{\textup{traffic}}^{l} \gets \textup{TVF\_Decoder}(P^l_{\textup{warp}},\hat{\textbf{p}}^l,\textup{TVF}^{l})$

  \State $(F^l,\mathbf{e}^l)\gets \textup{FlowHead}(\mathbf{e}_{\textup{point}}^{l},\textup{Interp}(\mathbf{e}^{l-1}),\mathbf{e}_{\textup{traffic}}^{l})$
\EndFor
\State \Comment{6. Static points correction}
\State $(F^L,\mathbf{S},\hat\Omega)\gets \textup{EgoHead}(F^L,\mathbf{e}^L,P^L)$
\State \Return $F^L,\mathbf{S},\hat\Omega$
\end{algorithmic}
\end{algorithm}

\subsection{Pseudo‑Code}

In this section, we present the pseudo‑code of TARS-ego (\cref{alg:main}) and detailed explanations of the TVF encoder and decoder modules (Algorithms \ref{alg:encoder} and \ref{alg:decoder}).

The forward pass of TARS‑ego takes $P, Q, O, h_{t-2}$ and outputs $F^L, \mathbf{S}, \hat{\Omega}, h_{t-1}$. It first computes low-level point features, updates temporal context with PointGRU \cite{pointgru}, and encodes point patch features using the MultiScaleEncoder. At the lowest level of TARS, the FlowHead initializes scene flow and embeddings using only point-level motion cues. A coarse‑to‑fine architecture then refines the flow: the TVF encoder builds the traffic-level motion understanding, the TVF decoder captures rigid motion in surrounding space, and the FlowHead produces refined flow from dual-level flow embeddings. Finally, the EgoHead corrects static points with predicted ego-motion.

\begin{algorithm}[h]
\caption{TVF\_Encoder}
\label{alg:encoder}
\begin{algorithmic}[1]
\Require 
  Previous points $P^{l-1}$, features $\mathbf{p}^{l-1}$, flow $F^{l-1}$, flow embeddings $\mathbf{e}^{l-1}$, $\textup{TVF}^{l-1}$, OD feature map $\chi^l_{\textup{od}}$
\Ensure Updated $\textup{TVF}^l$
\State \Comment{1. Scene update}
\If{$\textup{TVF}^{l-1}\neq\varnothing$ }   
  \State $X_{\textup{traffic}} \gets \textup{GRU}(\textup{Conv}(\chi^l_{\textup{od}}),\textup{TVF}^{l-1})$  \Comment{\cref{eq:GRU}}
\Else
  \State $X_{\textup{traffic}} \gets \textup{Conv}(\chi^l_{\textup{od}})$
\EndIf
\State \Comment{2. Flow painting}
\State $P^{l-1}_{\textup{warp}}\gets P^{l-1} + F^{l-1}$

\State $voxel \gets \textup{Voxelize2D}(P^{l-1}_{\textup{warp}},\textup{Concat}(\mathbf{p}^{l-1},\mathbf{e}^{l-1}))$  

\State $X_{\textup{motion}}\gets \textup{Point2Grid\_SelfAttn}(voxel)$

\State $X_{\textup{fusion}}\gets \textup{SpatialAttnFusion}(X_{\textup{traffic}},X_{\textup{motion}})$ \Comment{\cref{eq:fusion}}
\State \Comment{ Global attention}
\State $\textup{TVF}^{l}\gets \textup{AxialAttn}(X_{\textup{fusion}})$
\State \Return $\textup{TVF}^{l}$ 
\end{algorithmic}

\end{algorithm}

The Inputs to our TVF encoder (\cref{alg:encoder}) are previous level's $P^{l-1}$, $\mathbf{p}^{l-1}$, $F^{l-1}$, $\mathbf{e}^{l-1}$, $\mathrm{TVF}^{l-1}$, $\chi^l_{\mathrm{od}}$, and the output is the updated $\mathrm{TVF}^l$ for the current level. In the scene update stage, a ConvGRU \cite{ConvGRU} integrates prior TVF with OD features to encode traffic context. Then, in the flow painting stage, warped points are voxelized and processed by self‑attention to capture motion context. Finally, the spatial attention fuses traffic and motion streams, and the global attention enhances the combined representation.

\begin{algorithm}[h]
\caption{TVF\_Decoder}
\label{alg:decoder}
\begin{algorithmic}[1]
\Require 
  Warped points $P^l_{\textup{warp}}$, features $\hat{\textbf{p}}^l$, $\textup{TVF}^l$
\Ensure Traffic‑level flow embeddings $\mathbf{e}_{\textup{traffic}}^{l}$
\State $\mathcal{N}_{\textup{TVF}} \gets \textup{KNN}(K,P^l_{\textup{warp}},\textup{TVF}^l)$ \Comment{\cref{eq:decoder}}

\State $\mathbf{e}_{\textup{traffic}}^{l} \gets \textup{Grid2Point\_CrossAttn}(\hat{\textbf{p}}^l,\textup{TVF}^l,\textup{TVF}^l,\mathcal{N}_{\textup{TVF}})$ 
\State \Return $\mathbf{e}_{\textup{traffic}}^{l}$
\end{algorithmic}
\end{algorithm}

The TVF decoder (\cref{alg:decoder}) takes warped points $P^l_{\text{warp}}$, combined features $\hat{\mathbf{p}}^l$ and the $\text{TVF}^l$. It finds the $K$ nearest grid cells for each point and applies a grid‑to‑point cross‑attention. The output is the traffic‑level flow embedding $\mathbf{e}_{\text{traffic}}^l$, which is then consumed by the FlowHead.

\subsection{Reproduced LiDAR‑Model Details}

For a complete comparison with prior works, we include latest LiDAR scene flow models \cite{DeFlow, Flow4D} in the comparison on the VOD dataset, results shown in \cref{tab:VOD} in the main manuscript. In this section, we provide the details of our reproduction of DeFlow \cite{DeFlow} and Flow4D \cite{Flow4D}.

DeFlow and Flow4D underperform compared to TARS and CMFlow \cite{CMFlow}. This is because their fully-voxel representation, designed for the efficiency challenge in large-scale LiDAR point clouds, becomes unsuitable for sparse radar point clouds. In DeFlow \cite{DeFlow}, the GRU-based voxel-to-point refinement module becomes ineffective, because most pillars contain only a few radar point, leading to a loss of precise localization (although this effectively improves the efficiency). Similarly, Flow4D's \cite{Flow4D} fine-grain 4D voxel representation is less appropriate for sparse radar data.

Originally implemented on the LiDAR dataset Argoverse 2 \cite{argoverse}, DeFlow and Flow4D take two LiDAR point clouds $P \in \mathbb{R}^{N \times 3}$ and $Q \in \mathbb{R}^{M \times 3}$ as input. Note that, both of them take only $x,y,z$ coordinates as input channels, without using the intensity information from LiDAR. For fair comparison on the VOD dataset, we extend their input to 5 dimensions, adding RRV (relative radial velocity) and RCS (radar cross-section). We also adapt their grid feature map shape according to VOD's point cloud range.

Flow4D supports multi-frame input, while its 2-frame setup also achieves SOTA performance on Argoverse 2. As labeled in \cref{tab:VOD}, we test the Flow4D-2frame setup for a fair comparison with other models.

During training, we use the following losses: $\mathcal{L}_{\textup{sc}}$, $\mathcal{L}_{\textup{ss}}$, $\mathcal{L}_{\textup{rd}}$, $\mathcal{L}_{\textup{fg}}$, and $\mathcal{L}_{\textup{bg}}$ (details see \cref{appx:loss}). We keep their original implementations, and since these models lack ego-motion heads, we omit TARS-ego \& CMFlow’s additional losses ( $\mathcal{L}_{\textup{seg}}$, $\mathcal{L}_{\textup{ego}}$, $\mathcal{L}_{\textup{opt}}$). This setup is the same as the training of our TARS-no-ego model, where we also removed the ego-motion and motion-segmentation head (results shown in \cref{tab:ablation_vod} Group No.~1, highlighted in blue). Certainly, ego-motion compensation is also not applied for TARS-no-ego. Under the same training loss and ``no-ego'' setup, DeFlow and Flow4D still have a large accuracy gap compared to TARS-no-ego (16.7\% and 18.5\% on AccS; 27.7\% and 33.1\% on AccR, respectively).

\subsection{Dual‑Task Network Training Strategy} \label{appx:joint}

TARS jointly performs object detection and radar scene flow, enhancing the scene flow accuracy and also enabling a comprehensive perception. Experiments in PillarFlowNet \cite{PillarFlowNet} show that joint training of these two tasks reduces the performance of each individual task. Our experimental results of joint training are consistent with theirs, shown in \cref{tab3:joint}. Therefore, we perform a staged training: first train an object detector, then freeze its parameters to stabilize feature maps for the scene flow branch. This maintains OD performance while also allowing flexibility to adopt different training strategies for the object detector.

\begin{table}[htbp]
    \centering
    \caption{Joint training experiment on the proprietary dataset.}
    \label{tab3:joint}
    \resizebox{0.48\textwidth}{!}{
\begin{tabular}{ccc|cccc|cccc}
\hline
\hline
\multicolumn{3}{c|}{Training Strategy}             & \multicolumn{4}{c|}{Scene Flow Branch}                                 & \multicolumn{4}{c}{OD Branch AP {[}\%{]}$\uparrow$}                                                                \\ \hline
\multicolumn{1}{c|}{OD pre-trained} & Joint & Staged & MEPE$\downarrow$           & AccS$\uparrow$          & AccR$\uparrow$          & SEPE$\downarrow$           & Car           & Ped.          & Cycl.          & Truck         \\ \hline
\multicolumn{1}{c|}{no}             & \ding{51}     &        & 0.094          & 60.4          & 77.0          & \textbf{0.026} & 65.3          & 51.7          & 55.4          & 52.3          \\
\multicolumn{1}{c|}{yes, not frozen}            & \ding{51}     &        & \textbf{0.067} & 66.8          & \textbf{87.1} & 0.036          & 66.7          & 53.2          & 60.0          & 53.8          \\
\multicolumn{1}{c|}{yes,  frozen}            &       & \ding{51}      & 0.069          & \textbf{69.8} & 86.8          & 0.038          & \textbf{67.5} & \textbf{54.5} & \textbf{62.2} & \textbf{55.9} \\ \hline
\hline
\end{tabular}
}
\end{table}

\subsection{Object Detection Branch Details}  \label{appx:od_branch}

We use an off-the-shelf object detector as the OD branch on both datasets and freeze its weights. In this way, we maintain the OD performance and stabilize the feature map. As discussed in \cref{sec:3.1}, the object detection branch should have a grid-based backbone to provide bird’s eye view (BEV) feature maps to the scene flow branch. This excludes point-based detectors, such as 3D-SSD \cite{3D-SSD}, where features are preserved in point form rather than being structured into grids. In contrast, voxel-based detectors typically generate multi-scale feature maps, which aligns well with TARS's hierarchical architecture. Therefore, we utilize the OD feature map at each corresponding level within our hierarchical design.

We employ PointPillars \cite{PointPillars} as the object detector on the VOD dataset (result shown in \cref{tab:od_vod}). The OD branch result on the proprietary dataset is shown in \cref{tab:od_proprietary}.

\begin{table}[htbp]
 \centering
    \caption{Performance of the object detection branch on the VoD dataset. The results are reported separately for the entire annotated area and the driving corridor area.}
    \label{tab:od_vod}
    \resizebox{0.48\textwidth}{!}{
    \large
\begin{tabular}{c|cccc|cccc}
\hline
\hline
\multirow{2}{*}{Dataset} & \multicolumn{4}{c|}{AP in the entire annotated area [\%]$\uparrow$} & \multicolumn{4}{c}{AP in the driving corridor area [\%]$\uparrow$} \\ \cline{2-9} 
                         & Car      & Pedestrian    & \multicolumn{1}{c|}{Cyclist}   & mAP     & Car      & Pedestrian   & \multicolumn{1}{c|}{Cyclist}   & mAP     \\ \hline
VOD                      & 30.59    & 30.21         & \multicolumn{1}{c|}{61.95}     & 40.92   & 64.19    & 41.61        & \multicolumn{1}{c|}{85.04}     & 63.61   \\ \hline
\hline
\end{tabular}
}
\end{table}

\begin{table}[htbp]
 \centering
    \caption{Performance of the OD branch on the proprietary dataset.}
    \label{tab:od_proprietary}
    \resizebox{0.30\textwidth}{!}{
    \large
\begin{tabular}{c|cccc}
\hline
\hline
\multirow{2}{*}{Dataset} & \multicolumn{4}{c}{AP in the entire annotated area [\%]$\uparrow$} \\ \cline{2-5} 
                         & Car        & Pedestrian      & Cyclist      & Truck     \\ \hline
proprietary              & 67.46      & 54.50           & 62.15     & 55.86     \\ \hline
\hline
\end{tabular}
}
\end{table}

\section{Additional Ablation Studies}

\subsection{OD Branch Ablation Study} \label{appx:od_ablation}
In \cref{tab:od_ablation}, we test the effect of the object detection branch's accuracy on the performance of the scene flow branch. By adjusting the feature channels of PointPillars, we create three models of varying sizes: PP-L, PP-M, and PP-S (from large to small).

However, we observed that even with massive reductions in feature channels and parameters, PP-M and PP-S still maintain OD accuracy comparable to PP-L, and their corresponding scene flow branch performance is also similar. To isolate the corrective effect of the ego-motion head on scene flow predictions, we further test the performance of TARS-no-ego under three PointPillars setups. The experiments show that using PP-L achieves the best performance in both TARS-ego and TARS-no-ego models. However, due to the similar accuracy of the OD branch, the performance differences in the scene flow branch are not significant.

\begin{table}[htbp]
 \centering
    \caption{Ablation study of the OD branch on the VOD dataset. PP-S, PP-M, PP-L refer to three different PointPillars \cite{PointPillars} setups. TARS-no-ego: without ego-motion head and three supervision signals $\mathcal{L}_{\{ \textup{seg, ego, opt} \}}$. }
    \label{tab:od_ablation}
    \resizebox{0.48\textwidth}{!}{
    \large
\begin{tabular}{cccccc||cccccc}
\hline
\hline
\multicolumn{6}{c||}{Object Detection branch}                                                                                                                                          & \multicolumn{6}{c}{Scene Flow Branch}                                                                                \\ \hline
\multicolumn{1}{c|}{\multirow{2}{*}{Model}} & \multicolumn{1}{c|}{\multirow{2}{*}{\#Params}} & \multicolumn{4}{c||}{AP in the entire annotated area}                                   & \multicolumn{3}{c|}{TARS-ego}                                       & \multicolumn{3}{c}{TARS-no-ego}                \\ \cline{3-12} 
\multicolumn{1}{c|}{}                       & \multicolumn{1}{c|}{}                          & Car            & Ped.     & \multicolumn{1}{c|}{Cycl.}        & mAP            & EPE            & AccR          & \multicolumn{1}{c|}{AccS}          & EPE            & AccR          & AccS          \\ \hline
\multicolumn{1}{l|}{PP-S}                   & \multicolumn{1}{c|}{\textbf{0.05M}}            & 25.63          & 28.64          & \multicolumn{1}{c|}{61.04}          & 38.44          & 0.093          & 38.1          & \multicolumn{1}{c|}{68.5}          & 0.117          & 27.0          & 57.4          \\
\multicolumn{1}{l|}{PP-M}                   & \multicolumn{1}{c|}{0.33M}                     & 29.73          & 27.28          & \multicolumn{1}{c|}{59.33}          & 38.78          & 0.095          & 37.1          & \multicolumn{1}{c|}{67.2}          & 0.114          & 27.9          & 58.3          \\
\multicolumn{1}{l|}{PP-L}                   & \multicolumn{1}{c|}{9.30M}                     & \textbf{30.59} & \textbf{30.21} & \multicolumn{1}{c|}{\textbf{61.95}} & \textbf{40.92} & \textbf{0.092} & \textbf{39.0} & \multicolumn{1}{c|}{\textbf{69.1}} & \textbf{0.111} & \textbf{28.5} & \textbf{59.3} \\ \hline
\hline
\end{tabular}
}
\end{table}

\subsection{Class-wise Scene Flow Evaluation}

\citet{TrackFlow} proposed a novel scene flow evaluation method, computing class-aware EPE to analyze the failure cases in scene flow. We provide this analysis in \cref{tab2:deflow} for the best three models in the VOD dataset, i.e. DeFlow \cite{DeFlow},  CMFlow \cite{CMFlow} and our TARS-ego. Note that, Here we adapt the Bucket Normalized EPE \cite{TrackFlow} to MRNE. The reason is that, on the VOD dataset we compute MRNE for dynamic points, while EPE is an overall metric for all points. Therefore, to achieve class-aware evaluation, we compute per-class MRNE.

\begin{table}[htbp]
    \centering
    \caption{Scene flow evaluation on the VOD dataset, with per-class MRNE analysis. ``Sup.'': supervision.}
    \label{tab2:deflow}
    \resizebox{0.48\textwidth}{!}{
\begin{tabular}{lc|cccc|c|c||ccc}
\hline
\hline
                                                                    & \multicolumn{1}{l|}{} & \multicolumn{4}{c|}{Overall}                                        & Moving           & Static           & \multicolumn{3}{c}{Per-class MRNE$\downarrow$} \\ \hline
\multicolumn{1}{c|}{Method}                                         & Sup.                  & EPE$\downarrow$ & AccS$\uparrow$ & AccR$\uparrow$ & RNE$\downarrow$ & MRNE$\downarrow$ & SRNE$\downarrow$ & Car        & Ped.      & Cycl.     \\ \hline
\multicolumn{1}{l|}{DeFlow \cite{DeFlow}} & Cross                 & 0.217           & 11.8           & 31.6           & 0.087           & 0.098            & 0.085            & 0.092      & 0.081     & 0.106     \\
\multicolumn{1}{l|}{CMFlow \cite{CMFlow}} & Cross                 & 0.130           & 22.8           & 53.9           & 0.052           & 0.072            & 0.049            & 0.064      & 0.062     & 0.086     \\
\multicolumn{1}{l|}{TARS-ego (ours)}                                & Cross                 & \textbf{0.092}           & \textbf{39.0}           & \textbf{69.1}           & \textbf{0.037}           & \textbf{0.061}            & \textbf{0.034}            & \textbf{0.051}      & \textbf{0.052}     & \textbf{0.073}     \\ \hline
\hline
\end{tabular}
}
\end{table}

\begin{table}[htbp]
 \centering
    \caption{Comparison with fully-supervised models on the VOD dataset. Fully-supervised model results are cited from \cite{CMFlow}. ``Sup.'' indicates the supervision signal, Full: training with actual scene flow ground truth, Self: training with only self-supervised losses, Cross: with additional cross-modal losses.}
    \label{tab:sup_methods}
    \resizebox{0.48\textwidth}{!}{
    \large
\begin{tabular}{l|c|cccc}
\hline
\hline
Method                        & Sup. & EPE [m]$\downarrow$ & AccS [\%]$\uparrow$ & AccR [\%]$\uparrow$ & RNE [m]$\downarrow$ \\ \hline
FlowStep3D \cite{FlowStep3D}           & Full                 & 0.286             & 6.1            & 18.5           & 0.115             \\
Bi-PointFlowNet \cite{Bi-PointFlowNet} & Full                 & 0.242             & 16.4           & 35.0           & 0.097             \\
FlowNet3D \cite{FlowNet3d}             & Full                 & 0.201             & 16.9           & 37.9           & 0.081             \\
PointPWC-Net \cite{PointPWC}        & Full                 & 0.196             & 17.7           & 39.7           & 0.079             \\
PV-RAFT \cite{PV-RAFT}                 & Full                 & \textbf{0.126}    & \textbf{25.8}  & \textbf{58.7}  & \textbf{0.051}    \\ \hline
PointPWC-Net \cite{PointPWC}        & Self                 & 0.422             & 2.6            & 11.3           & 0.169             \\
FlowStep3D \cite{FlowStep3D}            & Self                 & 0.292             & 3.4            & 16.1           & 0.117             \\
CMFlow \cite{CMFlow}                   & Cross                & \textbf{0.141}    & \textbf{22.8}  & \textbf{53.9}  & \textbf{0.052}    \\ \hline
TARS-ego (ours)                               & Cross                & \textbf{0.092}    & \textbf{39.0}  & \textbf{69.1}  & \textbf{0.037}    \\ \hline
\hline
\end{tabular}
}
\end{table}

\subsection{Comparison with Fully-supervised Methods}

\citet{CMFlow} compared weakly-supervised CMFlow with fully-supervised LiDAR models, \eg PV-RAFT \cite{PV-RAFT}. We continue this discussion in \cref{tab:sup_methods}. Note that the ``fully-supervised" setup differs primarily in its annotations \cite{CMFlow}: using human-annotated bboxes and tracking IDs to derive actual scene flow GT. In contrast, the aforementioned pseudo scene flow GT $\hat{F}_{\text{fg}}$ in the foreground loss is generated using an off-the-shelf, pre-trained LiDAR multi-object tracking model, requiring no additional annotation efforts.

Experiments in \cite{CMFlow} demonstrated that the weakly-supervised CMFlow (with cross-modal losses) achieved $3\%$ lower AccS and $4.8\%$ lower AccR compared to the fully-supervised PV-RAFT. Nevertheless, CMFlow was able to achieve comparable EPE and AccR to the fully-supervised PV-RAFT by leveraging extra weakly-supervised training samples without costly annotations ($\sim140\%$ more than the number of training samples used for PV-RAFT). In contrast, as shown in \cref{tab:sup_methods}, TARS-ego surpasses fully-supervised PV-RAFT with a $13.2\%$ higher AccS and $10.4\%$ higher AccR, without requiring any additional training samples.

\begin{table}[htbp]
\centering
    \caption{TARS under fully-supervised setup on VOD.}
    \label{tab:full_sup}
    \resizebox{0.48\textwidth}{!}{
    \large
\begin{tabular}{l|c|cccc|c|c}
\hline
\hline
Method   & Sup.  & EPE$\downarrow$ & AccS$\uparrow$ & AccR$\uparrow$ & RNE$\downarrow$ & MRNE$\downarrow$ & SRNE$\downarrow$ \\
\hline
CMFlow \cite{CMFlow}   & Cross & 0.130           & 22.8           & 53.9           & 0.052           & 0.072            & 0.049            \\
PV-RAFT \cite{PV-RAFT} & Full  & 0.126           & 25.8           & 58.7           & 0.051           & N/A              & N/A              \\ \hline
TARS-ego (ours) & Cross & 0.092           & 39.0           & 69.1           & 0.037           & 0.061            & 0.034            \\
TARS-ego (ours) & Full  & \textbf{0.087}  & \textbf{42.2}  & \textbf{72.7}  & \textbf{0.035}  & \textbf{0.059}   & \textbf{0.031}   \\ \hline
\hline
\end{tabular}
}
\end{table}

\noindent\textbf{Training TARS under fully-supervised setup.} We test the performance of TARS-ego under the aforesaid fully-supervised setup and disabled the self-supervised losses: $\mathcal{L}_{\textup{sc}}, \mathcal{L}_{\textup{ss}}, \mathcal{L}_{\textup{rd}}, \mathcal{L}_{\textup{opt}}$. As shown in \cref{tab:full_sup}, the fully-supervised training improves the performance of TARS across all metrics, demonstrating TARS’s potential with more precise GT.

\subsection{Emergency‑Scenario Analysis}

\begin{figure*}[htbp]
    \centering
    
    % Top labels
    \makebox[0.3\linewidth]{Image Reference}%
    \makebox[0.3\linewidth]{~~~~~HALFlow}%
    \makebox[0.3\linewidth]{~~~~~TARS (ours)}%
    
    \vspace{2pt} % Adjust spacing between labels and images
    
    % Row 1
    \begin{subfigure}[b]{0.318\linewidth}
        \includegraphics[width=\linewidth]{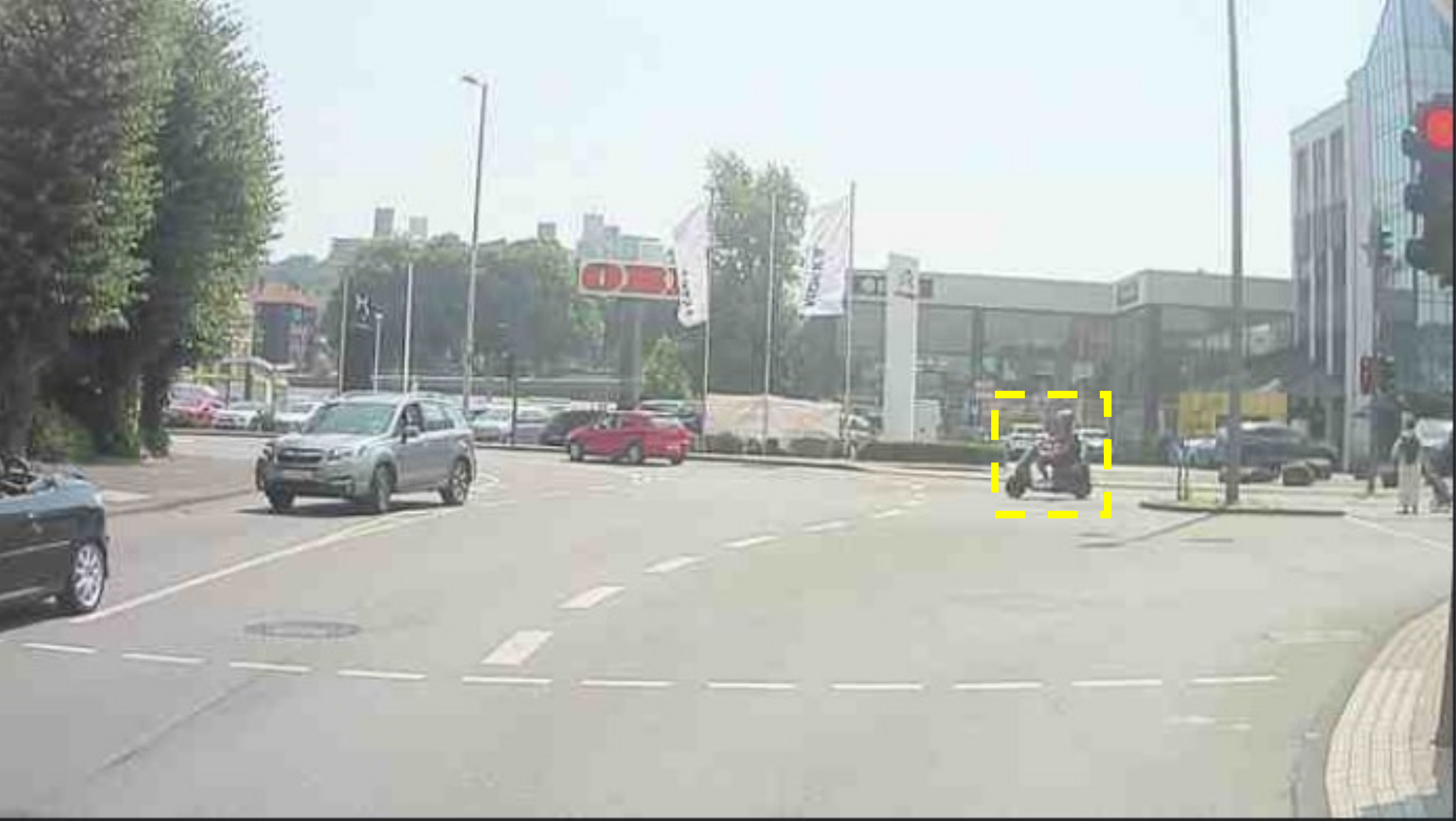}
    \end{subfigure}
    \hspace{-1pt}
    \begin{subfigure}[b]{0.3\linewidth}
        \includegraphics[width=\linewidth]{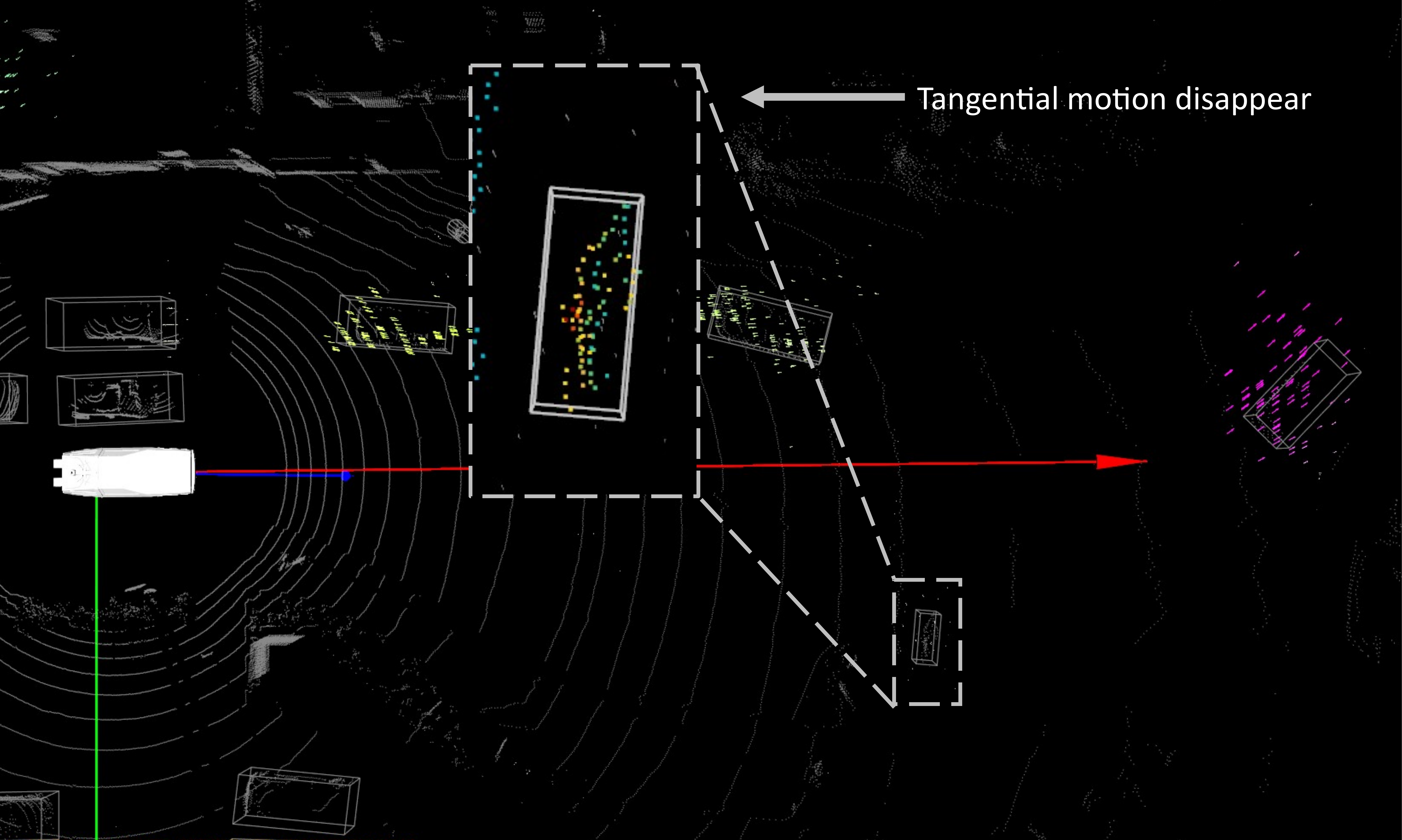}
    \end{subfigure}
    \hspace{-1pt}
    \begin{subfigure}[b]{0.3\linewidth}
        \includegraphics[width=\linewidth]{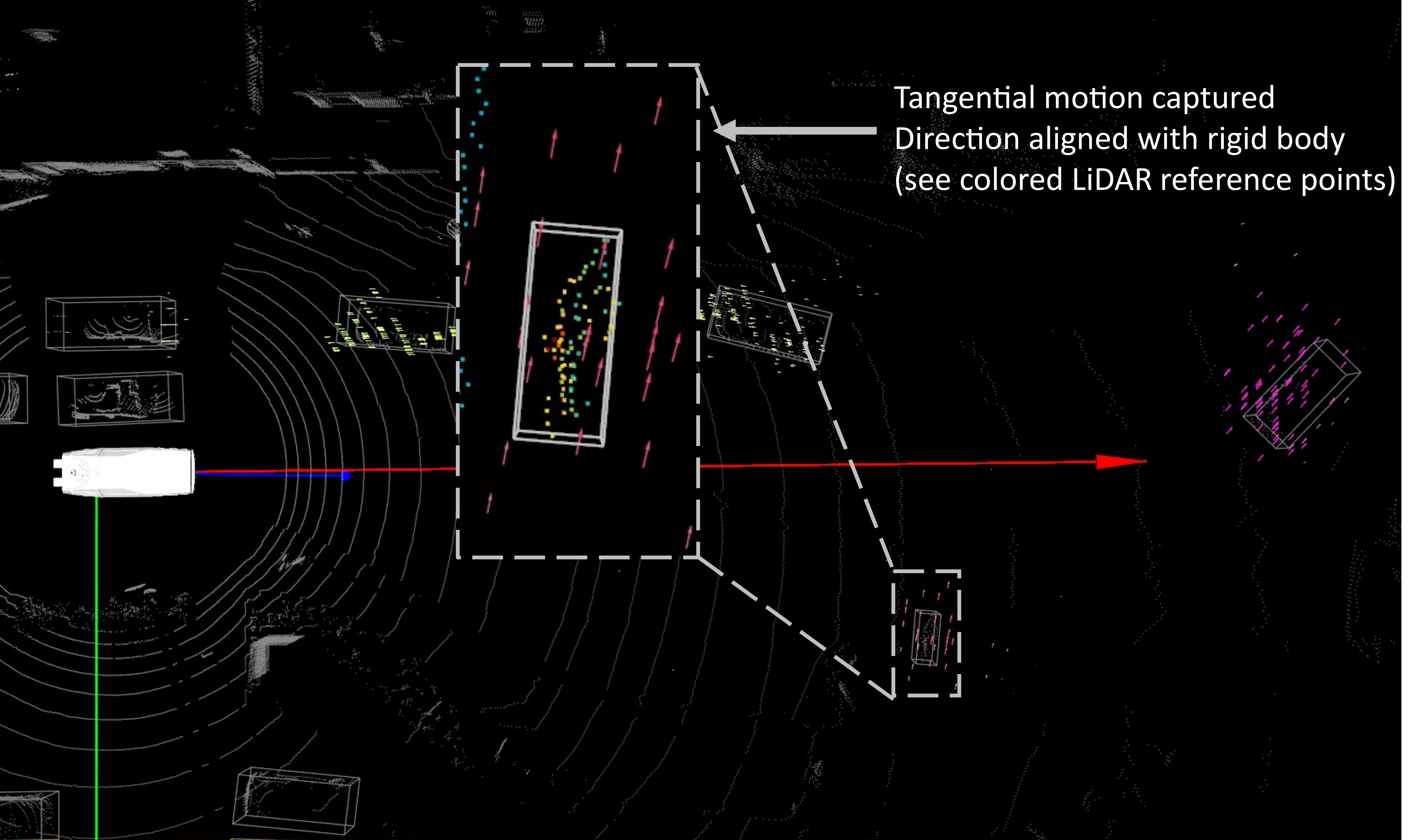}
    \end{subfigure}
    
    \vspace{2pt} % Adjust spacing between rows
    
    % Row 2
    \begin{subfigure}[b]{0.318\linewidth}
        \includegraphics[width=\linewidth]{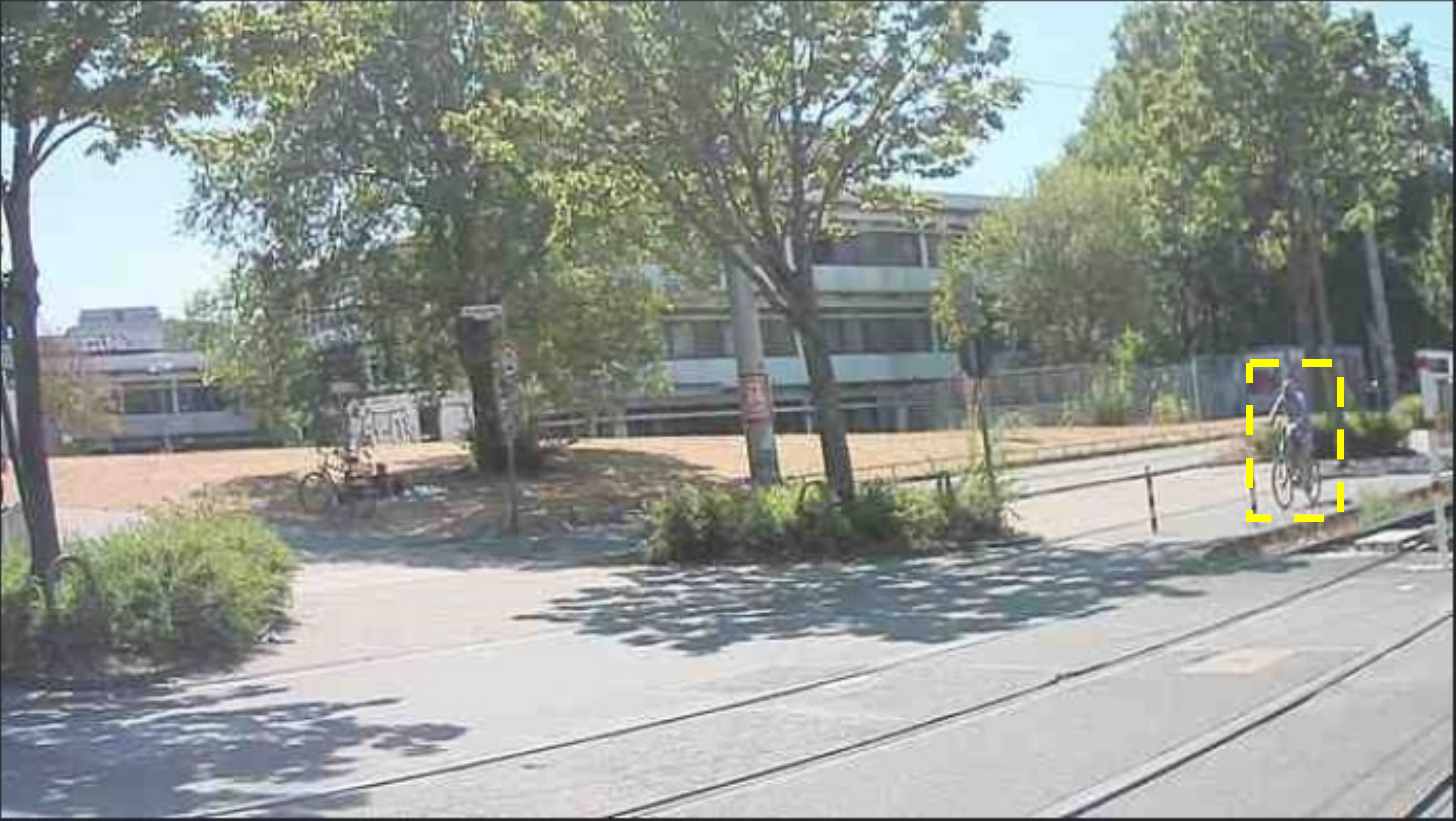}
    \end{subfigure}
    \hspace{-1pt}
    \begin{subfigure}[b]{0.3\linewidth}
        \includegraphics[width=\linewidth]{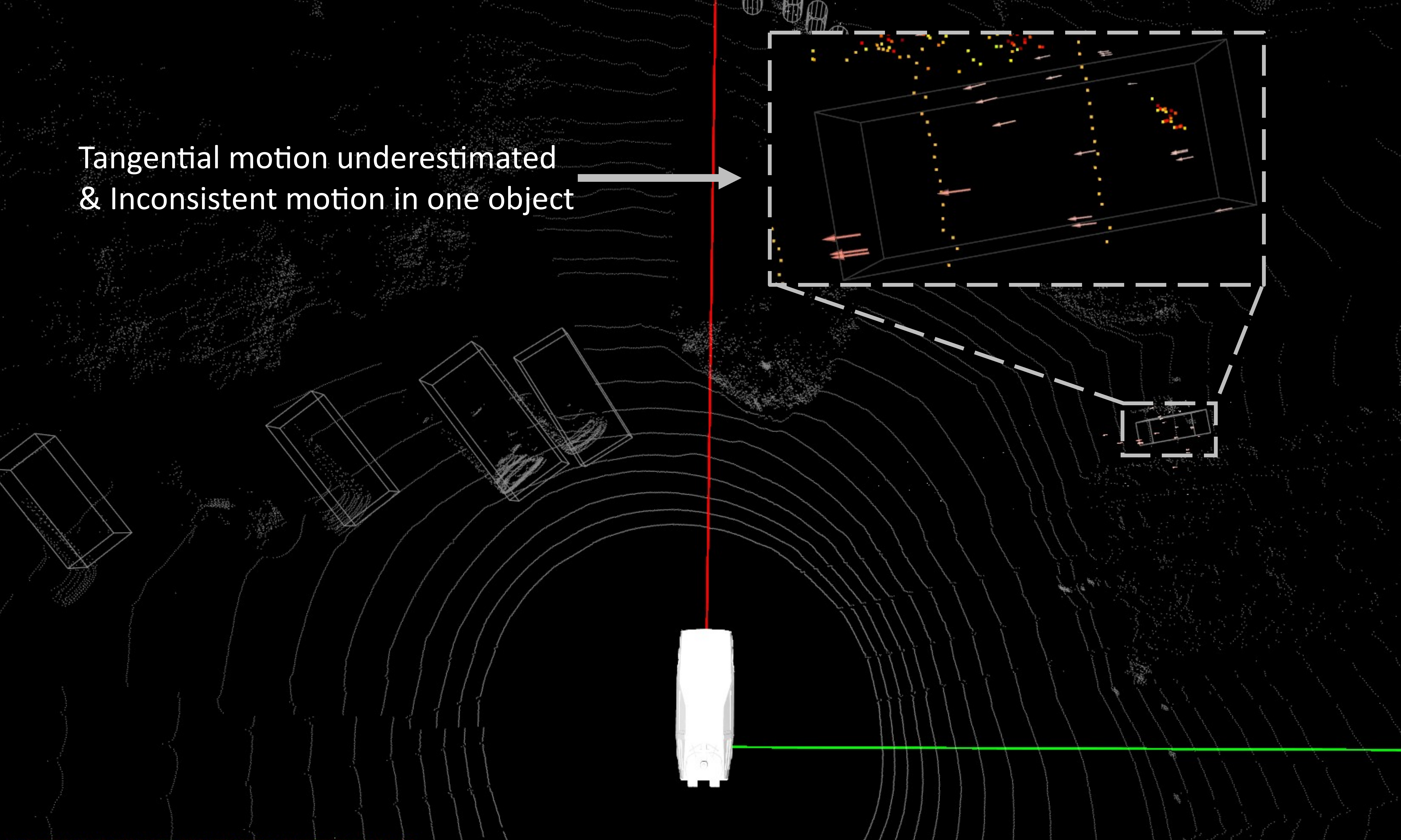}
    \end{subfigure}
    \hspace{-1pt}
    \begin{subfigure}[b]{0.3\linewidth}
        \includegraphics[width=\linewidth]{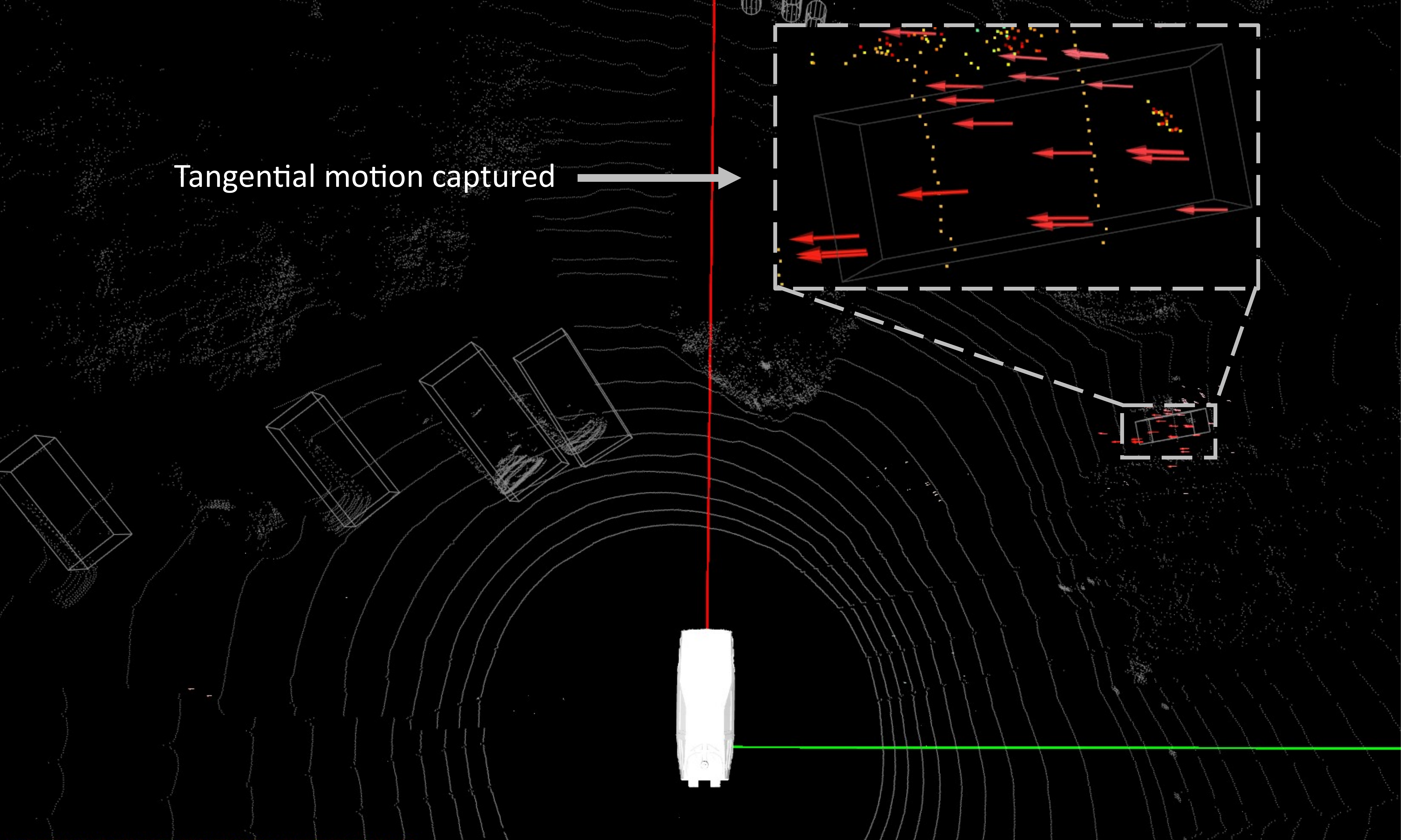}
    \end{subfigure}

    \caption{Qualitative results on the proprietary dataset, compared with HALFlow \cite{HALFlow}. It illustrates two scenarios of vulnerable road users approaching the front area of the ego vehicle with tangential motion. TARS is able to capture these challenging tangential movements.} \label{fig:quality_supp}
\end{figure*}

In TARS, perceiving traffic is one of the core ideas. Incorporating traffic modeling into motion prediction may raise concerns about emergent situations, such as a pedestrian suddenly entering traffic. Nevertheless, this is addressed by the overall design of TARS: all traffic-related modules in TARS employ attentive layers, enabling the neural network to adaptively balance the reliance on point-level and traffic-level features.

This is particularly evident in the encoding and decoding processes of the TVF. In the TVF encoder, we treat the point property $\textbf{p}^{l-1}$ (from PointNet) and motion features $\textbf{e}^{l-1}$ (flow embeddings) as equally important by concatenating them and then applying point-to-grid self-attention. Similarly, in the TVF decoder, we form the query by concatenating $\textbf{p}^{l}$ with the upsampled $\textbf{e}^{l-1}$ for grid-to-point cross-attention with the TVF.

More importantly, we leverage additional information from the OD branch, which provides object features of vulnerable road users. This helps the scene flow branch in perceiving the presence of vulnerable road users.

In \cref{fig:quality_supp}, we show two scenarios where a person riding a bike approaches the front area of the ego vehicle from a tangential direction, which is particularly challenging to perceive by radar sensors. Our TARS successfully captures the tangential motion of the bike, while, in the outputs of the previous work, the tangential motion erroneously disappears or is underestimated. 

These examples also highlights the importance of combining radar scene flow with object detection in autonomous driving. By integrating motion prediction with object bounding boxes, the system can select the optimal strategy based on the ego vehicle's speed and the direction and magnitude of target's motion.

\section{Efficiency Analysis} \label{appx:runtime}

We compare the efficiency of TARS with the previous best models on both datasets. All experiments are conducted on a single GeForce RTX 3090 GPU with a batch size of 1.

As shown in \cref{tab:runtime_vod}, on the VOD dataset, our scene flow branch has 3.82M parameters, which is fewer than CMFlow's 4.23M. The total runtime per frame of TARS-ego is 84ms, 15ms higher than CMFlow's 69ms, with 10ms attributed to the scene flow branch and 5ms to the additional OD branch. Nevertheless, TARS can achieve real-time performance under a 10-Hz radar, functioning as a unified system for both object detection and scene flow estimation.

The results on the proprietary dataset are shown in \cref{tab:runtime_prop}. Compared to the previous best model, i.e., HALFlow, TARS adds 0.6M additional parameters. The total runtime of TARS is 102ms per frame, 25ms higher than HALFlow's 77ms, with 19ms from the scene flow branch and 6ms from the additional OD branch. On the proprietary dataset, TARS operates at the edge of the real-time criterion and could achieve a real-time dual-task system with slight runtime optimization (\eg using FP16).

TARS has a higher runtime on the proprietary dataset compared to the VOD dataset. This results from the significantly larger number of radar points per frame ($\sim$6K vs. 256 points), and scene flow estimation operates on two input point clouds at the same time. On the VOD dataset, the increase in parameters is mainly due to the CNN layers for adapting the OD features in the TVF encoder.

\begin{table}[tbp]
 \centering
    \caption{Efficiency analysis on the VOD dataset. The total runtime is a sum of scene flow branch (SF) and OD branch (if exist). \#Param: the number of parameters in the scene flow branch. }
    \label{tab:runtime_vod}
    \resizebox{0.48\textwidth}{!}{
    \large
\begin{tabular}{l|ccc|ccc|c}
\hline
\hline
\multicolumn{1}{c|}{\multirow{2}{*}{Method}} & \multicolumn{3}{c|}{Performance (VOD)}                                & \multicolumn{3}{c|}{Runtime [ms]$\downarrow$} & \multirow{2}{*}{\#Param $\downarrow$} \\ \cline{2-7}
\multicolumn{1}{c|}{}                        & EPE [m]$\downarrow$ & AccS [\%]$\uparrow$ & AccR [\%]$\uparrow$ & Total           & SF           & OD           &                                           \\ \hline
CMFlow \cite{CMFlow}                       & 0.130               & 22.8                & 53.9                & 69              & 69           & /            & 4.23M                                     \\
TARS-ego (ours)                                  & 0.092               & 39.0                & 69.1                & 84              & 79           & 5            & 3.82M                                     \\ \hline
\hline
\end{tabular}
}
\end{table}

\begin{table}[tbp]
 \centering
    \caption{Efficiency analysis on the proprietary dataset. The total runtime is a sum of scene flow branch (SF) and OD branch.}
    \label{tab:runtime_prop}
    \resizebox{0.48\textwidth}{!}{
    \large
\begin{tabular}{l|ccc|ccc|c}
\hline
\hline
\multicolumn{1}{c|}{\multirow{2}{*}{Method}} & \multicolumn{3}{c|}{Performance (proprietary)} & \multicolumn{3}{c|}{Runtime [ms]$\downarrow$} & \multirow{2}{*}{\#Param $\downarrow$} \\ \cline{2-7}
\multicolumn{1}{c|}{}                        & MEPE [m]$\downarrow$       & AccS [\%]$\uparrow$     & AccR [\%]$\uparrow$     & Total       & SF      & OD      &                         \\ \hline
HALFlow \cite{HALFlow}                                      & 0.170      & 50.9     & 63.8     & 77          & 77        & /       & 1.00M                   \\
TARS (ours)                                         & 0.069      & 69.8     & 86.8     & 102         & 96        & 6       & 1.60M                   \\ \hline
\hline
\end{tabular}
}
\end{table}

\section{Evaluation Metrics \& Loss Functions}

\subsection{Evaluation Metrics} \label{appx:metrics}
In this section, we introduce details of the evaluation metrics used in the experiments. Our input consists of two point clouds $P$ and $Q$. Let $\hat{\mathbf{f}}_i$ and $\mathbf{f}_i$ represent the GT and predicted scene flow for a point $p_i$, respectively. $N_{\text{fg}}$ and $N_{\text{bg}}$ denote the number of moving points and static points. The subscripts $\text{fg}$ or $\text{bg}$ indicate points belonging to moving or static point set.

On the VOD dataset, we follow the evaluation metrics used in CMFlow \cite{CMFlow}:
\begin{itemize}
    \item $\text{EPE} = \frac{1}{N} \sum_{i=1}^{N} \|\mathbf{f}_i - \hat{\mathbf{f}}_i\|_2,$
    \item $\text{AccS} = \frac{1}{N} \sum_{i=1}^{N} \mathbb{I}(\text{EPE}_i < 0.05),$
    \item $\text{AccR} = \frac{1}{N} \sum_{i=1}^{N} \mathbb{I}(\text{EPE}_i < 0.1),$
    \item $\text{RNE} = \frac{1}{N} \sum_{i=1}^{N} \text{EPE}_i / \frac{{r_R}_i}{ {r_L}_i},$ RNE normalizes EPE by the resolution ratio between radar and LiDAR at each point location (independent of model predictions), in order to accommodate low-resolution radar. The average resolution ratio on VOD is 2.5. For high-resolution radar datasets, we directly report EPE \& MEPE.
    \item $\text{MRNE} = \frac{1}{N_{\text{fg}}} \sum_{i \in P_\text{fg}} \text{RNE}_i,$
    \item $\text{SRNE} = \frac{1}{N_{\text{bg}}} \sum_{i \in P_\text{bg}} \text{RNE}_i,$
\end{itemize}
where $\mathbb{I}(\cdot)$ is the indicator function, which evaluates to one if and only if the condition is true, $\frac{r_R}{r_L}$ is the resolution ratio between radar and LiDAR.

On the proprietary dataset, high-resolution radars eliminate the need for RNE, and we focus on the evaluation of moving points since ego-motion compensation is applied:
\begin{itemize}
    \item $\text{MEPE} = \frac{1}{N_{\text{fg}}} \sum_{i \in P_\text{fg}} \|\mathbf{f}_i - \hat{\mathbf{f}}_i\|_2,$
    \item $\text{MagE} = \frac{1}{N_\text{fg}} \sum_{i \in P_\text{fg}} \left| \|\mathbf{f}_i\|_2 - \|\hat{\mathbf{f}}_i\|_2 \right|,$
    \item $\text{DirE} = \frac{1}{N_\text{fg}} \sum_{i \in P_\text{fg}} \arccos\left(\frac{\mathbf{f}^T_i \cdot \hat{\mathbf{f}}_i}{\|\mathbf{f}_i\|_2 \|\hat{\mathbf{f}}_i\|_2}\right),$
    \item $\text{AccS} = \frac{1}{N_{\text{fg}}} \sum_{i \in P_\text{fg}} \mathbb{I}(\text{MEPE}_i < 0.05),$
    \item $\text{AccR} = \frac{1}{N_{\text{fg}}} \sum_{i \in P_\text{fg}} \mathbb{I}(\text{MEPE}_i < 0.1),$
    \item $\text{SEPE} = \frac{1}{N_{\text{bg}}} \sum_{i \in P_\text{bg}} \|\mathbf{f}_i - \hat{\mathbf{f}}_i\|_2,$
    \item $\text{AvgEPE} = \frac{1}{2} (\text{MEPE} + \text{SEPE}),$
\end{itemize}
where AccS and AccR are computed only on moving points. MagE and DirE have not appeared in prior works. They reflect the two aspects of EPE: magnitude error and directional error. They can help reveal the sources of MEPE, \eg high-speed underestimation or direction error.

\begin{table*}[t]
    \centering
    \caption{SOTA object detection (OD) and tracking accuracy on the nuScenes and VOD dataset.}
    \label{tab1:radarOD}
    \resizebox{0.7\textwidth}{!}{
\begin{tabular}{c|cc|cc|ccc}
\hline
\hline
\multirow{2}{*}{Sensor} & \multicolumn{2}{c|}{nuScenes SOTA OD [\%]$\uparrow$} & \multicolumn{2}{c|}{VOD SOTA OD [\%]$\uparrow$}        & \multicolumn{3}{c}{SOTA Tracking [\%]$\uparrow$}                                        \\ \cline{2-8} 
                        & \multicolumn{1}{c|}{Method}                  & mAP            & \multicolumn{1}{c|}{Method}                     & mAP           & \multicolumn{1}{c|}{Method}             & \multicolumn{1}{c|}{Dataset}  & AMOTA         \\ \hline
LiDAR                   & \multicolumn{1}{c|}{LION (NeurIPS24) \cite{LION}}        & \textbf{69.8}  & \multicolumn{1}{c|}{CMFA (ICRA24) \cite{ODbase}} & \textbf{69.6} & \multicolumn{1}{c|}{VoxelNeXt (CVPR23) \cite{VoxelNeXt}} & \multicolumn{1}{c|}{nuScenes} & \textbf{71.0} \\ \hline
Radar                   & \multicolumn{1}{c|}{RadarDistill (CVPR24) \cite{RadarDistill}}   & 20.5           & \multicolumn{1}{c|}{CMFA (ICRA24) \cite{ODbase}} & 41.8          & \multicolumn{1}{c|}{RaTrack (ICRA24) \cite{RaTrack}}   & \multicolumn{1}{c|}{VOD}      & 31.5          \\ \hline
\hline
\end{tabular}
}
\vspace{-0.3cm}
\end{table*}

\subsection{Weakly-Supervised Loss Functions} \label{appx:loss}
In our weakly-supervised training, we employ the three self-supervised losses proposed in \cite{RaFlow}: the soft Chamfer loss $\mathcal{L}_{\textup{sc}}$, spatial smoothness loss $\mathcal{L}_{\textup{ss}}$, and radial displacement loss $\mathcal{L}_{\textup{rd}}$ . We also use the foreground loss $\mathcal{L}_{\textup{fg}}$ \cite{CMFlow} and our background loss $\mathcal{L}_{\textup{bg}}$.

On the VOD dataset, we train TARS-ego with additional cross-modal losses from CMFlow \cite{CMFlow}: the motion segmentation loss $\mathcal{L}_{\textup{seg}}$, ego-motion loss $\mathcal{L}_{\textup{ego}}$, optical flow loss $\mathcal{L}_{\textup{opt}}$. TARS-superego and TARS-no-ego didn't use these losses.

Let $P'=P_{\textup{warp}}$ denote the resulting point cloud that warped by the predicted scene flow.

\noindent \textbullet \enspace The soft Chamfer loss $\mathcal{L}_{\textup{sc}}$ minimizes distances between nearest points between $P_{\textup{warp}}$ and $Q$ while handling outliers using probabilistic matching, formulated as:
\begin{align} \label{eq1}
\mathcal{L}_{\textup{sc}} = & \sum_{p'_i \in P_{\textup{warp}}} \mathbb{I}(\nu(p'_i) > \delta) 
\left[\min_{q_j \in Q} \|p'_i - q_j\|_2^2 - \epsilon \right]_+ \nonumber \\
+ & \sum_{q_i \in Q} \mathbb{I}(\nu(q_i) > \delta) 
\left[\min_{p'_j \in P_{\textup{warp}}} \|q_i - p'_j\|_2^2 - \epsilon \right]_+,
\end{align}
where $\nu(p)$ represents the per-point Gaussian density factor estimated using kernel density estimation, points with $\nu(p)$ below threshold $\delta$ are discarded as outliers, and the application of $[\cdot]_+ = \max(0, \cdot)$ in \cref{eq1} ensures that small matching discrepancies below $\epsilon$ are ignored. For further details, we refer to RaFlow \cite{RaFlow}.

\noindent \textbullet \enspace The spatial smoothness loss $\mathcal{L}_{\textup{ss}}$ enforces neighboring points to have similar flow vectors, weighted by distance to ensure spatial smoothness, formulated as:
\begin{equation}
\mathcal{L}_{\text{ss}} = \sum_{p_i \in P} \sum_{p_j \in \mathcal{N}_P(p_i)} k(p_i, p_j) \|\mathbf{f}_i - \mathbf{f}_j\|_2^2,
\end{equation}
where $k(p_i, p_j) = \exp\left(-\frac{\|p_i - p_j\|_2^2}{\alpha}\right)$ is a radial basis function (RBF) kernel that weighs each neighbor point $p_j \in \mathcal{N}_P(p_i)$ based on its Euclidean distance to $p_i$, with $\alpha$ controlling the impact of the distance. All kernel weight values are normalized together using a softmax function.

\noindent \textbullet \enspace The radial displacement loss $\mathcal{L}_{\textup{rd}}$ constrains the radial projection of predicted flow vectors using RRV measurements, formulated as:
\begin{equation}
\mathcal{L}_{\text{rd}} = \sum_{p_i \in P} \left|  \frac{ \mathbf{f}^T_i \cdot p_i}{\|p_i\|} - \textup{RRV}_{i} \Delta t \right|,
\end{equation}
where $p_i$ denotes the 3D point coordinate, $\textup{RRV}_{i}$ is the RRV measurement of a point, and $\Delta t$ is the time interval between two radar scans.

\noindent \textbullet \enspace The foreground loss $\mathcal{L}^l_{\textup{fg}}$ uses pseudo scene flow GT $\hat{F}_{\textup{fg}}$ derived from the object bounding box and tracking ID generated by an off-the-shelf LiDAR multi-object tracking model. This loss is applied for each level and formulated as:
\begin{equation}
\mathcal{L}^l_{\textup{fg}} = \frac{1}{N^l_{\textup{fg}}} \sum\nolimits_{i=1}^{N^l_{\textup{fg}}} \left\| \hat{\mathbf{f}}^{l}_{\textup{fg}_i} - \mathbf{f}^{l}_{\textup{fg}_i}  \right\|_2,
\end{equation}
where $\hat{\mathbf{f}}^{l}_{\textup{fg}_i}$ is the pseudo GT for $i$-th moving point.

\noindent \textbullet \enspace The background loss $\mathcal{L}^{l}_{\textup{bg}}$ uses pseudo GT $\hat{F}_{\textup{bg}}$ derived from ego-motion or using zero vectors if ego-motion compensated, applied for level $l$ and formulated as:
\begin{equation}
\mathcal{L}^l_{\textup{bg}} = \frac{1}{N^l_{\textup{bg}}} \sum\nolimits_{i=1}^{N^l_{\textup{bg}}} \left\| \hat{\mathbf{f}}^{l}_{\textup{bg}_i} - \mathbf{f}^{l}_{\textup{bg}_i}  \right\|_2,
\end{equation}
where $\hat{\mathbf{f}}^{l}_{\textup{bg}_i}$ is the pseudo GT for $i$-th static point.

\noindent \textbullet \enspace The overall loss $\mathcal{L}_{\textup{all}}$ is a sum of the above losses:
\begin{equation}
     \mathcal{L}_{\textup{all}} =\mathcal{L}_{\textup{sc}} + \mathcal{L}_{\textup{ss}} + \mathcal{L}_{\textup{rd}} + \sum\nolimits_{l=1}^{L}(\mathcal{L}^l_{\textup{fg}} + \lambda_{\textup{bg}}\mathcal{L}^l_{\textup{bg}}).
\end{equation}

On the VOD dataset, we use additional cross-modal losses \cite{CMFlow}: the motion segmentation loss $\mathcal{L}_{\textup{seg}}$, ego-motion loss $\mathcal{L}_{\textup{ego}}$, optical flow loss $\mathcal{L}_{\textup{opt}}$.

\noindent \textbullet \enspace The motion segmentation loss $\mathcal{L}_{\text{seg}}$ uses the pseudo segmentation GT derived from the odometer and RRV to train a motion-segmentation head, formulated as:
\begin{equation}
\mathcal{L}_{\text{seg}} = \frac{1}{2} \sum_{i=1}^{N}\hat{s}_i\log(s_i) + (1-\hat{s}_i)\log(1-s_i),
\end{equation}
where $\hat{s}_i \in \{0, 1\}$ is the pseudo motion segmentation GT derived from odometer and RRV, and $s_i \in [0, 1]$ is the predicted moving probability.

\noindent \textbullet \enspace The ego-motion loss $\mathcal{L}_{\text{ego}}$ uses the GT ego-motion from the odometer to train an ego-motion head (\cref{appx:ego_head}), formulated as:
\begin{equation}
\mathcal{L}_{\text{ego}} = \frac{1}{N} \sum_{i=1}^{N} \left\| \left( \Omega - \Omega_{\textup{pred}} \right) 
\begin{bmatrix}
p_i \\ 
1
\end{bmatrix}
\right\|_2,
\end{equation}
where $\Omega_{\textup{pred}}$ is the predicted rigid transformation, $\Omega$ is the GT ego-motion derived from odometry.

\noindent \textbullet \enspace The optical flow loss $\mathcal{L}_{\text{opt}}$ projecting the scene flow onto an image and training with pseudo optical flow labels from an off-the-shelf optical flow model as additional supervision signal, formulated as:
\begin{equation}
\mathcal{L}_{\text{opt}} = \frac{1}{N_{\textup{fg}}} \sum_{i=1}^{N_{\textup{fg}}} \, D_\textup{Point-Ray} \left(p_i + \textbf{f}_i, \textup{Ray}(p^{\textup{proj}}_i + \widehat{\textbf{f}_i^\textup{opt}}), \theta\right),
\end{equation}
where $p^{\textup{proj}}_i$ is the projection of point $p_i$ on the image plane, $\widehat{\textbf{f}_i^\textup{opt}}$ is pseudo optical flow GT generated by a off-the-shelf image optical flow model, and $D_\textup{Point-Ray}(\cdot)$ calculates the point-line distance between the warped 3D point and the corresponding $\textup{Ray}(\cdot)$ traced from the optical flow-warped pixel. The parameter $\theta$ denotes the sensor calibration parameters. This loss is computed only for moving points. For further details, we refer to \cite{CMFlow}.

\noindent \textbullet \enspace The overall loss on the VOD dataset is formulated as: 
\begin{align}
     \mathcal{L}_{\textup{all}} = & \mathcal{L}_{\textup{sc}} + \mathcal{L}_{\textup{ss}} + \mathcal{L}_{\textup{rd}} + \sum\nolimits_{l=1}^{L}(\mathcal{L}^l_{\textup{fg}} + \lambda_{\textup{bg}}\mathcal{L}^l_{\textup{bg}}) \nonumber \\
+ & \mathcal{L}_{\textup{seg}} + \mathcal{L}_{\textup{ego}} + \lambda_{\textup{opt}}\mathcal{L}_{\textup{opt}},
\end{align}
where $\lambda_{\textup{opt}}$ is set to 0.1 as CMFlow \cite{CMFlow}.

\section{Discussion on the Joint Network}

TrackFlow \cite{TrackFlow} directly derives LiDAR scene flow from object detection and tracking results. This approach is intuitive, especially as it directly uses the object-detection bounding boxes to ensure motion rigidity. However, this method is unsuitable for radar, where sparsity and noise cause a 30-50 mAP drop in object detection (OD) and a 40 AMOTA gap in tracking (see \cref{tab1:radarOD}). This leads to missing motion predictions for false negatives and frequent tracking failures, which severely limit its ability to infer radar scene flow. By contrast, our point-wise radar scene flow predicts motion per point, ensuring stable motion signals. Even if objects are undetected or noisily tracked, some moving points can still be identified, providing auxiliary information for decision-making in autonomous driving. In our design, TARS perceives traffic context by leveraging OD feature maps instead of bounding boxes, thereby reducing reliance on high OD recall.

    \small
    \putbib    % 输出附录引用
\end{bibunit}

\end{document}